\documentclass[10pt,twocolumn]{article} 

\usepackage{OptimLabTwoColumn}

\usepackage{mathtools}
\usepackage{algorithm}
\usepackage{algpseudocode}
\usepackage{amsmath,amssymb,amsthm,amsfonts}
\usepackage{booktabs}
\usepackage{subfigure}
\usepackage{url}
\usepackage{makecell}
\usepackage{multirow}
\usepackage[symbol]{footmisc}
\usepackage{tikz}
\usepackage{caption}
\usepackage{hyperref}
\usepackage{bbm}
\usepackage{kotex}

\usepackage{bbding} % checkmark
\usepackage{pifont} % checkmark

\usepackage{algorithm} 
\usepackage{algpseudocode} 

\usetikzlibrary{shapes,decorations,arrows,calc,arrows.meta,fit,positioning}
\tikzset{
    -Latex,auto,node distance =1 cm and 1 cm,semithick,
    state/.style ={ellipse, draw, minimum width = 0.7 cm},
    point/.style = {circle, draw, inner sep=0.04cm,fill,node contents={}},
    bidirected/.style={Latex-Latex,dashed},
    el/.style = {inner sep=2pt, align=left, sloped}
}

\def\cL{{\cal L}}

\def\cL{{\cal L}}

\newcommand{\bZ}{{\bf Z}}
\newcommand{\bz}{{\bf z}}

\newcommand{\bX}{{\bf X}}
\newcommand{\bx}{{\bf x}}

\newcommand{\mbR}{\mathbb{R}}
\newcommand{\mbE}{\mathbb{E}}

\newcommand{\bc}{\begin{center}}
\newcommand{\ec}{\end{center}}
\newcommand{\be}{\begin{equation}}
\newcommand{\ee}{\end{equation}}
\newcommand{\ba}{\begin{array}}
\newcommand{\ea}{\end{array}}
\newcommand{\bean}{\begin{eqnarray*}}
\newcommand{\eean}{\end{eqnarray*}}
\newcommand{\bea}{\begin{eqnarray}}
\newcommand{\eea}{\end{eqnarray}}
\newcommand{\ben}{\begin{enumerate}}
\newcommand{\een}{\end{enumerate}}
\newcommand{\bed}{\begin{itemize}}
\newcommand{\eed}{\end{itemize}}

\newtheorem{definition}{Definition}
\newtheorem{assumption}{Assumption}

\begin{document}

\title{Balanced Marginal and Joint Distributional Learning\\via Mixture Cramer-Wold Distance}
% High-dimensional Multiple-Categorical Synthetic Data Generation via Cramer-Wold Distance

\author{
  Seunghwan An, Sungchul Hong, \textnormal{and} Jong-June Jeon\thanks{Corresponding author.} \\
  % Department of Statistics, University of Seoul, S. Korea \\
  Department of Statistical Data Science, University of Seoul, S. Korea \\
  \texttt{\{dkstmdghks79, shong, jj.jeon\}@uos.ac.kr} \\
}

\maketitle
\thispagestyle{empty}

\begin{abstract}
In the process of training a generative model, it becomes essential to measure the discrepancy between two high-dimensional probability distributions: the generative distribution and the ground-truth distribution of the observed dataset. Recently, there has been growing interest in an approach that involves slicing high-dimensional distributions, with the Cramer-Wold distance emerging as a promising method. However, we have identified that the Cramer-Wold distance primarily focuses on joint distributional learning, whereas understanding marginal distributional patterns is crucial for effective synthetic data generation. In this paper, we introduce a novel measure of dissimilarity, the mixture Cramer-Wold distance. This measure enables us to capture both marginal and joint distributional information simultaneously, as it incorporates a mixture measure with point masses on standard basis vectors. Building upon the mixture Cramer-Wold distance, we propose a new generative model called CWDAE (Cramer-Wold Distributional AutoEncoder), which shows remarkable performance in generating synthetic data when applied to real tabular datasets. Furthermore, our model offers the flexibility to adjust the level of data privacy with ease.
\end{abstract}

\section{Introduction}
\label{sec:1}

Training a generative model requires measuring the discrepancy between two high-dimensional probability distributions: the generative distribution and the ground-truth distribution of the observed dataset. 
% Then, the generative model can produce a synthetic dataset resembling the original dataset. 
However, this task demands substantial computational resources and introduces growing sample complexity \cite{Genevay2018SampleCO}. 
% To address these challenges, prominent approaches have introduced the concept of a low-dimensional latent space, upon which a conditional generative model is defined.

The Variational Autoencoder (VAE) framework \cite{Kingma2014, JimenezRezende2014StochasticBA} and the Generative Adversarial Network (GAN) framework \cite{Goodfellow2014GenerativeAN, larsen2016autobeyond, Munjal2019ImplicitDI}, have been well-established approaches to address these challenges. VAE incorporates the conditional independence assumption, resulting in an element-wise reconstruction loss that often produces blurry images \cite{larsen2016autobeyond}. On the other hand, GAN employs an adversarial loss, but it can lead to the \textit{mode collapse} problem, where the generative model fails to capture the full data variability \cite{Heusel2017GANsTB}.

More recent and promising approaches \cite{Kolouri2018SlicedWA, Deshpande2018GenerativeMU, Knop2020GenerativeMW} involve comparing two high-dimensional distributions directly by slicing high-dimensional distributions over their one-dimensional marginals and comparing their marginal distributions \cite{Kolouri2022GeneralizedSP}. Especially, \cite{Tabor2018CramerWoldA} proposed the Cramer-Wold distance, which can be seen as a combination of the sliced approach with the MMD distance and can be effectively computed due to the closed-form formula. And, in the domain of high-dimensional image datasets, \cite{Knop2020GenerativeMW} applied the Cramer-Wold distance as a measure for the discrepancy and proposed a generative model, CW2, which tackled the issue of mode collapse. Also, this approach eliminated the requirement for adversarial training and showed enhanced performance in synthetic data generation.

However, our empirical examination has revealed that the generative model proposed by \cite{an2023distributional}, referred to as DistVAE, demonstrates competitive joint distributional similarity metric scores to CW2 despite relying solely on element-wise reconstruction loss. This observation suggests that the acquisition of marginal distributional knowledge plays a significant role in enhancing the performance of synthetic data generation. This occurrence can be attributed to the fact that the point masses on standard basis vectors are zero, rendering it hard to calculate the discrepancy between two distributions on an element-wise level using the Cramer-Wold distance-based reconstruction loss.

Our primary contribution is that we introduce a novel generative model, which is designed to strike a balance between learning the marginal and joint distributions effectively. We achieve this balance by devising a reconstruction loss that is based on a mixture measure comprising point masses on standard basis vectors and a normalized surface measure, and the mixture measure is employed for the integral measure of the Cramer-Wold distance. Furthermore, our proposed reconstruction loss still maintains a closed-form solution, which facilitates its practical implementation.

To showcase the effectiveness of our model in accurately capturing the underlying distribution of datasets, particularly in terms of both marginal and joint distributional similarity, we conduct an evaluation of our generative model using real tabular datasets for the purpose of synthetic data generation. This evaluation serves as a demonstration of the practical applicability and performance of our proposed approach.

\section{Related Work}
\label{sec:2}

\textbf{Distributional learning (generative model learning).}
Distributional learning (generative model learning) involves estimating the underlying distribution of an observed dataset. Generative models based on latent spaces aim to perform distributional learning by generating data closely resembling a given dataset. An early and prominent example of generative modeling is the VAE. However, VAE faced limitations in generative performance due to a misalignment between the distribution of representations that learned information from the observations, known as the aggregated posterior, and the prior distribution.

To address this issue, \cite{Makhzani2015AdversarialA} introduced the Adversarial AutoEncoder (AAE), which directly minimizes the aggregated posterior and prior using the adversarial loss from the GAN framework \cite{Goodfellow2014GenerativeAN}. This introduction of adversarial loss made it easier to compute, even when the aggregated posterior took complex forms, unlike KL-divergence. Similarly, \cite{Bousquet2017FromOT} proposed the penalized optimal transport (POT). The POT's objective function consists of a reconstruction loss (cost) and a penalty term that minimizes the divergence (distance) between the aggregated posterior and prior distributions. Subsequent research incorporated various divergences into this penalty term, such as Maximum Mean Discrepancy (MMD) \cite{Tolstikhin2017WassersteinA}, Sliced-Wasserstein distance \cite{Deshpande2018GenerativeMU}, and the Cramer-Wold distance \cite{Tabor2018CramerWoldA}. Notably, Sliced-Wasserstein and the Cramer-Wold distance are based on random projections of high-dimensional datasets onto one-dimensional subspaces, resolving challenges in calculating distances between multivariate distributions.

\textbf{Modeling of the decoder and reconstruction loss.} 
To increase the distributional capacity, many papers have focused on decoder modeling while not losing the mathematical link to maximize the ELBO. \cite{Takahashi2018StudenttVA, akrami2022lesion} assume their decoder distributions as Student-$t$ and asymmetric Laplace distributions, respectively. \cite{Barron2017AGA} proposes a general distribution of the decoder, which allows improved robustness by optimizing the shape of the loss function during training. Recently, \cite{bredell2023explicitly} proposes a reconstruction loss that directly minimizes the \textit{blur error} of the VAE by modeling the covariance matrix of multivariate Gaussian decoder. More recently, \cite{an2023distributional} introduced the \textit{continuous ranked probability score} (CRPS) in the ELBO of the VAE framework, a proper scoring rule that measures the distance between the proposed cumulative distribution function (CDF) and the ground-truth CDF of the underlying distribution. It shows theoretically that it is feasible to minimize the KL-divergence between the ground-truth density and the density estimated through generative modeling. 

On the other hand, there exists a research direction that focuses on replacing the reconstruction loss without concern for losing the mathematical derivation of the lower bound. \cite{larsen2016autobeyond, Munjal2019ImplicitDI} replace the reconstruction loss with an adversarial loss of the GAN framework. \cite{Hou2016DeepFC} introduces a feature-based loss that is calculated with a pre-trained convolutional neural network (CNN). Another approach by \cite{czolbe2020watson} adopts Watson's perceptual model, and \cite{Jiang2020FocalFL} directly optimizes the generative model in the frequency domain by a focal frequency reconstruction loss. 
% Most of the above-mentioned methods aim to capture the properties of human perception by replacing the element-wise loss ($L_1$ or $L_2$-norm), which hinders the reconstruction of images \cite{larsen2016autobeyond}.
Furthermore, while many papers commonly use divergence for alignment in the latent space, some studies directly introduce divergence (or distance) into the data space. For instance, papers like \cite{Dziugaite2015TrainingGN, Li2015GenerativeMM} utilized Maximum Mean Discrepancy (MMD), and \cite{Deshpande2018GenerativeMU} employed the sliced Wasserstein distance for reconstruction error. Similarly, \cite{Knop2020GenerativeMW} introduced a kernel distance into the reconstruction loss based on the Cramer-Wold distance.

\textbf{Synthetic data generation.}
The synthetic data generation task actively adopts the GAN framework, as it allows for nonparametric synthetic data generation \cite{xu2019ctgan, zhao2021ctabgan, Fang2022OvercomingCO, li2021cdfgan, Hernandez2022SyntheticDG, Yale2020GenerationAE}. \cite{xu2019ctgan, zhao2021ctabgan} assume that continuous columns in tabular datasets can be approximated using Gaussian mixture distributions and model their decoder accordingly. They also employ the Variational Gaussian mixture model \cite{Blei2016VariationalIA}, known as \textit{mode-specific normalization}, to preprocess the continuous variables. However, this preprocessing step requires additional computational resources and hyperparameter tuning to determine the number of modes. 
% \cite{zhao2021ctabgan} focus on regularizing the difference between the first and second-order statistics of the observed and synthetic datasets. 
\cite{Fang2022OvercomingCO, li2021cdfgan} employed CDF to transform continuous columns into uniformly distributed variables to handle multi-modal and long tail continuous variables and alleviate the gradient vanishing problem.

On the other hand, several studies have focused on capturing the correlation structure between variables to improve the quality of synthetic data. For instance, \cite{Yang2019GroupedCG} maximizes the correlation between two different latent vectors representing diseases and drugs. Similarly, \cite{Kuo2022TheHG} introduces an alignment loss based on the $L_2$ distance between correlation matrices. On the other hand, \cite{Fang2022OvercomingCO} modifies the Multilayer Perceptron (MLP) with Convolutional Neural Networks (CNN).

\subsection{Notations}

Let $I = I_c \cup I_d = \{1, \cdots, p\}$ be an index set of the variables, where $I_c$ and $I_d$ are the index sets of continuous and discrete variables, respectively. $\bx_j \in \mbR$ for $j \in I_c$ and $\bx_j \in \{0, 1\}^{T_j}$ for $j \in I_d$, where $T_j$ denotes the number of levels, and it means that the discrete variables are transformed into a one-hot vector.
% where $\Delta^{T_j-1}$ is the standard $(T_j-1)$-simplex and $T_j$ denotes the number of levels. 

Let $\bx = (\bx_1,\cdots,\bx_p) \in \mbR^{D}$ be an observation consisting of continuous and one-hot encoded discrete variables and $D = |I_c| + \sum_{j \in I_d} T_j$. We denote the ground-truth underlying distribution (probability density function, PDF) as $p(\bx)$ and the ground-truth CDF (Cumulative Distribution Function) as $F(\bx)$.

Let $\bz$ be a latent variable, where $\bz \in \mbR^d$ and $d < D$. The prior distribution of $\bz$ are assumed to be $p(\bz) \coloneqq \mathcal{N}(\bz|\mathbf{0}, \mathbf{I})$ (standard Gaussian distribution), where $\mathbf{I}$ is $d \times d$ identity matrix. For an arbitrary posterior distribution $q(\bz|\bx;\phi)$, the aggregated posterior distribution \cite{Makhzani2015AdversarialA, tomczak2018vae} is defined as 
% $q(\bz;\phi) \coloneqq \int q(\bz|\bx;\phi) p(\bx) d\bx$ 
\bean
q(\bz;\phi) \coloneqq \int q(\bz|\bx;\phi) p(\bx) d\bx.
\eean
Furthermore, we define a generative model based on the aggregated posterior as follows:
\bean
\hat{p}(\bx;\theta,\phi) \coloneqq \int_{\mbR^d} p(\bx|\bz;\theta) q(\bz;\phi) d\bz.
% \hat{p}(\bx;\theta) &\coloneqq& \int_{\mbR^d} p(\bx|\bz;\theta) p(\bz) d\bz.
% &=& \int_{\mbR^d} \delta(\bx - G(\bz;\theta)) p(\bz) d\bz,
\eean

Lastly, let $\sigma_D$ be the normalized surface measure on $S_D$, where $S_D$ denotes the unit sphere in $\mbR^D$. For a sample $R = \{r_i\}_{i=1}^n \subset \mbR$, 
\bean
\mbox{sm}_{\gamma}(R) \coloneqq \frac{1}{n} \sum_{i=1}^n N(r_i, \gamma),
\eean
which is a smoothen distribution with a Gaussian kernel $N(\cdot, \gamma)$ and $N(m, s)$ denotes the one-dimensional normal density with mean $m$ and variance $s$.

\section{Proposal}
\label{sec:3}

Our primary goal is to improve the Cramer-Wold distance by integrating marginal distributional learning simultaneously. Our comparative analysis between CW2 \cite{Knop2020GenerativeMW} and DistVAE \cite{an2023distributional} prompted this initiative concerning the influence of marginal reconstruction losses on the performance of synthetic data generation, as elaborated in the subsequent section.

\subsection{Motivation}
\label{sec:3.1}

\textbf{Cramer-Wold distance.} 
\cite{Tabor2018CramerWoldA} made use of the Cramer-Wold Theorem \cite{cramer1939} and the Radon Transform \cite{deans2007radon} to simplify the computation of the distance between two multivariate distributions into one-dimensional calculations. Notably, the Cramer-Wold distance bears similarity to the sliced-Wasserstein distribution \cite{Kolouri2018SlicedWA, Deshpande2018GenerativeMU}. However, \cite{Tabor2018CramerWoldA} demonstrated that the Cramer-Wold distance can be computed in a closed form without the need for sampling, provided each multivariate distribution is smoothed with a Gaussian kernel. 

\begin{definition}[Cramer-Wold distance \cite{Tabor2018CramerWoldA}] \label{def:cw}
Let two PDFs $p(\bx)$ and $\hat{p}(\bx)$ are given, where $\bx \in \mbR^D$. The Cramer-Wold distance $d_{CW}^2$ with $\sigma_D$ is defined as
\bean
&& d_{CW}^2\Big(p(\bx), \hat{p}(\bx); \sigma_D\Big) \\
&\coloneqq& \int_{S_D} \left\| \mbox{sm}_\gamma(\nu^\top\bX) - \mbox{sm}_\gamma(\nu^\top\hat{\bX}) \right\|_2^2 d\sigma_D(\nu),
\eean
where $\nu \in S_D$, $\nu^\top\bX \coloneqq \{\nu^\top\bx^{(i)}\}_{i=1}^n$ for $\bx^{(i)} \sim p(\bx)$, and $\nu^\top\hat{\bX} \coloneqq \{\nu^\top\hat{\bx}^{(i)}\}_{i=1}^n$ for $\hat{\bx}^{(i)} \sim \hat{p}(\bx)$.
% $\bX_{\eta_1} \coloneqq \{\bx^{(i)}\}_{i=1}^n$, $\bx^{(i)} \sim p(\bx;\eta_1)$, $\bX_{\eta_2} \coloneqq \{\bx^{(i)}\}_{i=1}^n$, $\bx^{(i)} \sim p(\bx;\eta_2)$, 
\end{definition}

Noth that $\nu \in \mbR^p$ is a unit-norm projection vector such that $\|\nu\|_2 = 1$. In this paper, as per Definition \ref{def:cw}, the Cramer-Wold distance is assumed to be computed with finite $n$ samples. Additionally, in this section, we simplify notation by considering only continuous variables in explaining model descriptions and motivation without the loss of generality.

\textbf{CW2.}
In a related advancement, \cite{Knop2020GenerativeMW} introduced the AutoEncoder-based generative model CW2, which is based on the Cramer-Wold distance. CW2 is notable for being the first model effectively trained using a kernel distance on high-dimensional datasets. Unlike some other models, it does not require adversarial training and does not require sampling projection vectors (i.e., it has a closed-form solution). Additionally, \cite{Knop2020GenerativeMW} demonstrated that employing the Cramer-Wold distance as a reconstruction loss, especially in tasks like image generation, can significantly enhance the generative performance of the model. This approach also maintains the implicit correlation structure of multivariate datasets by directly comparing multivariate distributions.
% Inspired by the image generation performance of \cite{Knop2020GenerativeMW}, we have also incorporated the Cramer-Wold distance as a reconstruction loss.

The encoder $q(\bz|\bx;\phi)$ and decoder $p(\bx|\bz;\theta)$ of CW2 are assumed to be 
\bean
q(\bz|\bx;\phi) &=& \delta(\bz - \mu(\bx;\phi)) \\
p(\bx|\bz;\theta) &=& \delta(\bx - G(\bz;\theta)),
\eean
where $\delta$ is the Dirac delta function, and $\mu: \mbR^D \mapsto \mbR^d$ and $G: \mbR^d \mapsto \mbR^D$ are neural networks parameterized by $\phi$ and $\theta$, respectively. And the objective of CW2 is to minimize $\cL_{CW2}$ with respect to $\theta, \phi$, and $\cL_{CW2}$ is defined as 
\bean
\cL_{CW2}(\theta,\phi) &\coloneqq& d_{CW}^2\Big(p(\bx), \hat{p}(\bx;\theta,\phi); \sigma_D\Big) \\
&+& \lambda \cdot \log d_{CW}^2\Big(p(\bz), q(\bz;\phi); \sigma_d\Big),
\eean
where $\lambda > 0$.
% \bean
% \hat{p}(\bx;\theta,\phi) = \int_{\mbR^d} p(\bx|\bz;\theta) q(\bz;\phi) d\bz.
% % &=& \int_{\mbR^d} \delta(\bx - G(\bz;\theta)) p(\bz) d\bz,
% \eean

\textbf{DistVAE.}
\cite{an2023distributional} proposed a novel distributional learning method of VAE, called DistVAE, which aims to effectively capture the underlying distribution of the observed dataset using a nonparametric approach. \cite{an2023distributional} defined the probability model of the observation as an infinite mixture of asymmetric Laplace distribution (ALD), leading to a direct estimation of CDFs using the CRPS loss \cite{gasthaus2019probabilistic} while maintaining the mathematical derivation of the lower bound.

The objective of DistVAE is to minimize $\cL_{Dist}$ with respect to $\theta,\phi$, and $\cL_{Dist}$ is defined as 
\bean
&& \cL_{Dist}(\theta,\phi) \\
&\coloneqq& \mbE_{p(\bx)} \mbE_{q(\bz|\bx;\phi)} \left[ \sum_{j=1}^D \int_0^1 \rho_{\alpha}\Big( \bx_j - Q_j(\alpha|\bz; \theta_j) \Big) d\alpha \right] \nonumber\\
&+& \beta \cdot \mbE_{p(\bx)} [\mathcal{KL}(q(\bz|\bx;\phi) \| p(\bz))],
\eean
where $\theta = (\theta_1, \cdots, \theta_{D})$, $\beta$ is a non-trainable constant, $\rho_{v}(u) = u(v - \mathbb{I}(u < 0))$ (check function), and $\mathbb{I}(\cdot)$ denotes the indicator function. $Q_j(\cdot|\cdot; \theta_j): [0, 1] \times \mbR^d \mapsto \mbR$ is the location parameter of ALD, which is parameterized with $\theta_j$. We parameterize the function $Q_j$, the location parameter of ALD, by a linear isotonic spline (see \cite{gasthaus2019probabilistic, an2023distributional} for detailed parameterization and the closed form reconstruction loss).

\begin{assumption}[Smooth quantile function] \label{assump:Q}
Given an arbitrary $\theta_j$, $Q_j(\cdot|\bz; \theta_j)$ is invertible and differentiable for all $j$ and $\bz \in \mbR^d$.
\end{assumption}

The posterior $q(\bz|\bx;\phi)$ and the decoder $p(\bx|\bz;\theta)$ are assumed as
\bean
q(\bz|\bx;\phi) &=& \mathcal{N}\big(\bz | \mu(\bx;\phi), diag(\sigma^2(\bx;\phi))\big) \\
p(\bx|\bz;\theta) &=& \prod_{j=1}^D \frac{d}{d\bx_j} Q_j^{-1}(\bx_j|\bz;\theta_j),
\eean
under Assumption \ref{assump:Q}, where $\mu:\mbR^{D} \mapsto \mbR^d$, $\sigma^2:\mbR^{D} \mapsto \mbR_+^d$ are neural networks parameterized with $\phi$, and $diag(a), a \in \mbR^d$ denotes a diagonal matrix with diagonal elements $a$. It implies that the conditional quantile functions are employed to construct the decoder. 

% As introduced in \cite{gasthaus2019probabilistic, an2023distributional}, we parameterize the function $Q_j$, the location parameter of ALD, by a linear isotonic spline as follows:
% \bean \label{eq:isotonic}
% Q_j(\alpha|\bz;\theta_j) &=& \gamma^{(j)}(\bz) + \sum_{m=0}^M b_m^{(j)} (\bz) (\alpha - d_m)_+ \\
% &\mbox{\quad s.t. \quad}& \sum_{m=0}^k b_m^{(j)}(\bz) \geq 0, k=1,\cdots,M,
% \eean
% where $\gamma^{(j)}(\bz) \in \mbR$, $b^{(j)}(\bz) = (b_0^{(j)}(\bz), \cdots, b_M^{(j)}(\bz)) \in \mbR^{M+1}$, $d = (d_0, \cdots, d_M) \in [0, 1]^{M+1}$, $0 = d_0 < \cdots < d_M = 1$, and $(u)_+ \coloneqq \max(0, u)$. $\theta_j$ is a neural network parameterized mapping such that $\theta_j: \mbR^d \mapsto \mbR \times \mbR^{M+1}$, which takes $\bz$ as input and outputs $\gamma^{(j)}(\bz)$ and $b^{(j)}(\bz)$. The constraint is introduced to ensure monotonicity. 

\begin{table*}[ht]
\caption{Marginal and joint distributional similarity. $\uparrow$ denotes higher is better and $\downarrow$ denotes lower is better.}
  \centering
  % \resizebox{\textwidth}{!}{
  \begin{tabular}{lrrrrrrrrrrrrrrrr}
    \toprule
    & \multicolumn{2}{c}{marginal} & \multicolumn{4}{c}{joint} \\
    \cmidrule(lr){2-3} \cmidrule(lr){4-7}
    Model & KS $\downarrow$ & W1 $\downarrow$ & PCD $\downarrow$ & log-cluster $\downarrow$ & MAPE $\downarrow$ & F1 $\uparrow$  \\
    \midrule
CW2 & $0.081_{\pm 0.029}$ & $0.093_{\pm 0.025}$ & $1.348_{\pm 0.810}$ & $-3.983_{\pm 1.334}$ & $0.318_{\pm 0.344}$ & $0.644_{\pm 0.085}$ \\
DistVAE & $\textbf{0.056}_{\pm 0.025}$ & $\textbf{0.066}_{\pm 0.023}$ & $1.481_{\pm 0.386}$ & $-3.642_{\pm 0.751}$ & $0.301_{\pm 0.194}$ & $\textbf{0.670}_{\pm 0.074}$ \\
DistVAE (+CW) & $0.059_{\pm 0.022}$ & $0.071_{\pm 0.028}$ & $\textbf{0.972}_{\pm 0.331}$ & $\textbf{-4.828}_{\pm 1.349}$ & $\textbf{0.232}_{\pm 0.112}$ & $0.658_{\pm 0.080}$ \\
    \bottomrule
  \end{tabular}
  % }
\label{tab:motiv}
\end{table*}

\begin{figure*}
    \centering
    \subfigure[DistVAE]{
    \includegraphics[width=0.23\textwidth]{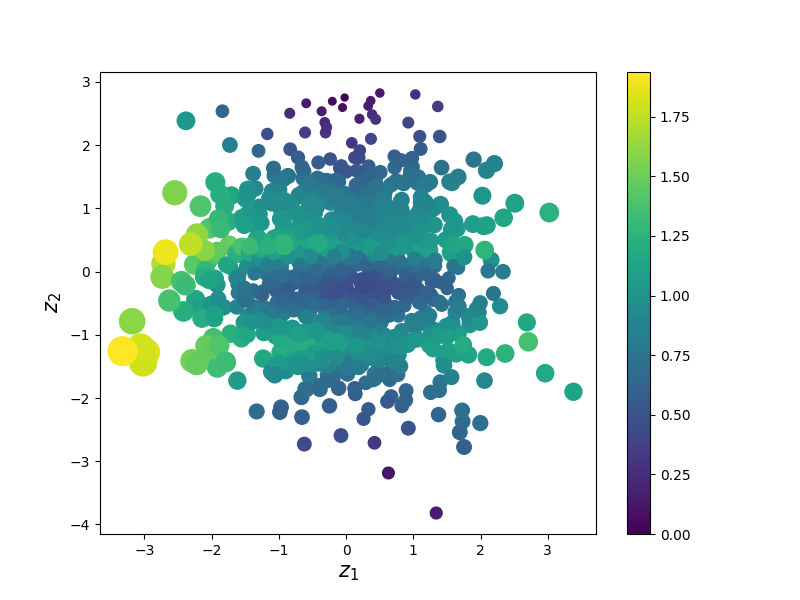}
    }
    \subfigure[DistVAE (+CW)]{
    \includegraphics[width=0.23\textwidth]{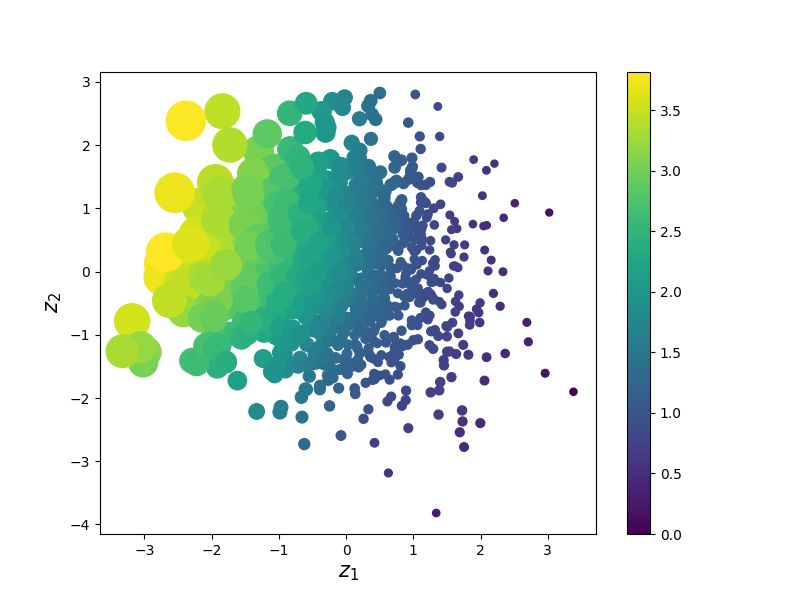}
    }
    \subfigure[DistVAE]{
    \includegraphics[width=0.23\textwidth]{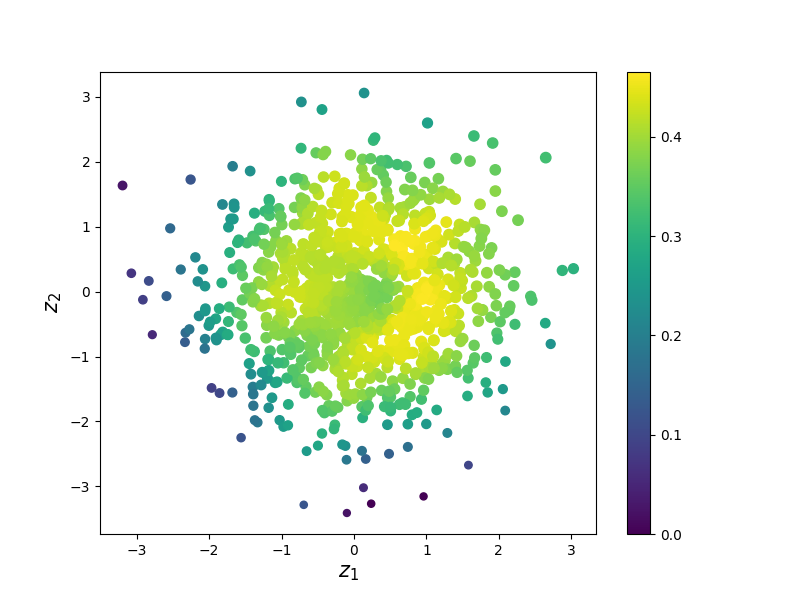}
    }
    \subfigure[DistVAE (+CW)]{
    \includegraphics[width=0.23\textwidth]{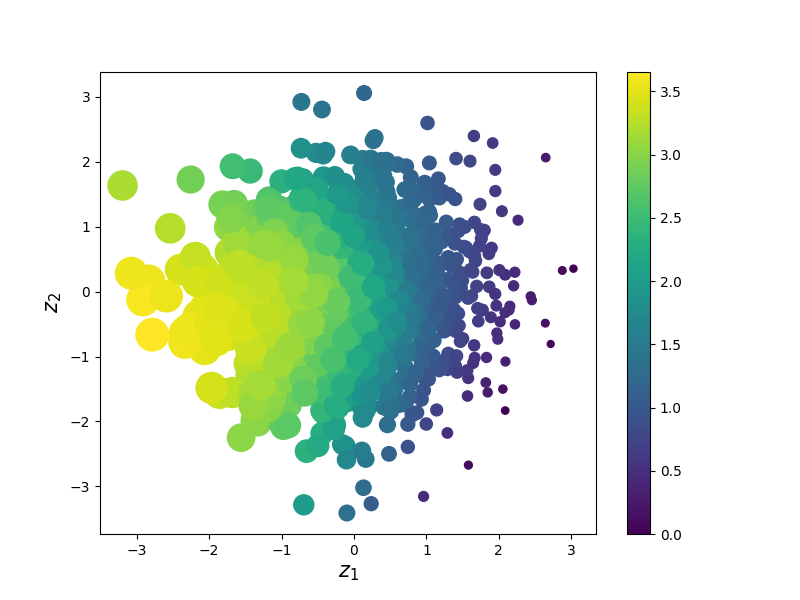}
    }
    % \subfigure[DistVAE, \texttt{cabs}. $\bx_1$: Life Style Index, $\bx_2$: Customer Rating.]{
    % \includegraphics[width=0.4\textwidth]{fig/latent/DistVAE_cabs_Life_Style_Index_Customer_Rating_latent.png}
    % }
    % \subfigure[DistVAE + CW, \texttt{cabs}. $\bx_1$: Life Style Index, $\bx_2$: Customer Rating.]{
    % \includegraphics[width=0.4\textwidth]{fig/latent/DistVAE_CW_cabs_Life_Style_Index_Customer_Rating_latent.png}
    % }
    % \subfigure[DistVAE, \texttt{kings}. $\bx_1$: lat, $\bx_2$: sqft living15.]{
    % \includegraphics[width=0.4\textwidth]{fig/latent/DistVAE_kings_lat_sqft_living15_latent.png}
    % }
    % \subfigure[DistVAE + CW, \texttt{kings}. $\bx_1$: lat, $\bx_2$: sqft living15.]{
    % \includegraphics[width=0.4\textwidth]{fig/latent/DistVAE_CW_kings_lat_sqft_living15_latent.png}
    % }
    \caption{Scatter plots of the latent variables and synthetic samples, which are generated from DistVAE and DistVAE (+CW).
    % \textcolor{blue}{위의 3가지 데이터셋은 PCD 측면에서 특히 Cramer-Wold distance를 사용하였을 때 성능향상이 많이 이루어짐.}
    }
    \label{fig:latent1}
\end{figure*}

\begin{assumption}[Conditional independence] \label{assump:indep}
$\bx_j, \forall j$, are conditionally independent given $\bz$. 
\end{assumption}

\textbf{Comparison and motivation.}
\cite{an2023distributional} demonstrated the capability of DistVAE to recover the ground-truth PDF in terms of the KL-divergence. However, it is crucial to emphasize that this recovery is feasible only when Assumption \ref{assump:indep} is satisfied. The limitation arises because \cite{an2023distributional} exclusively utilized marginal quantile functions, as modeling multivariate quantiles can be quite a challenging task \cite{Feldman2021CalibratedMQ}. 

When Assumption \ref{assump:indep} is violated, the following equations demonstrate the discrepancy:
\bean
q(\bx|\bz;\phi) \neq \prod_{j=1}^D q_j(\bx_j|\bz;\phi),
\eean
where 
\bean
q(\bx|\bz;\phi) &\coloneqq& \frac{p(\bx)q(\bz|\bx;\phi)}{q(\bz;\phi)} \\
q_j(\bx_j|\bz;\phi) &=& \int_{\mbR^{D-1}} q(\bx|\bz;\phi) d\bx_{-j},
\eean
and $\bx_{-j}$ denotes the vector of $\bx$ excluding $\bx_j$ for $j \in {1,\cdots,D}$. This violation leads to:
\bea \label{eq:ineq}
\hat{p}(\bx;\theta^*(\phi)) &=& \int_{\mbR^d} \prod_{j=1}^D \frac{d}{d\bx_j} Q_j^{-1}(\bx_j,\bz;\theta_j^*(\phi)) q(\bz;\phi) d\bz \nonumber\\
&=& \int_{\mbR^d} \prod_{j=1}^D q_j(\bx_j|\bz) q(\bz;\phi) d\bz \nonumber\\
&\neq& \int_{\mbR^d} q(\bx|\bz) q(\bz;\phi) d\bz \\
&=& \int_{\mbR^d} p(\bz) q(\bz|\bx;\phi) d\bz \nonumber\\
&=& p(\bx),\nonumber
\eea
where $\theta^*(\phi)$ is the optimal solution of the reconstruction loss of $\cL_{Dist}(\theta,\phi)$
% $\theta^*(\phi) = \arg\min_{\theta} \mbE_{p(\bx)} \mbE_{q(\bz|\bx;\phi)} \left[ \sum_{j=1}^D \int_0^1 \rho_{\alpha}\Big( \bx_j - Q_j(\alpha|\bz; \theta_j) \Big) d\alpha \right]$
% $\theta^*(\phi) = \arg\min_{\theta} \cL_{Dist}(\theta,\phi)$ 
for a fixed $\phi$.

Inspired by the image generation performance of \cite{Knop2020GenerativeMW} and the Cramer-Wold distance's capability to measure the discrepancy between two multivariate distributions directly, we incorporate the following Cramer-Wold distance as an additional regularization term into DistVAE:
\bea \label{eq:cwreg}
&& d_{CW}^2\Big(p(\bx), \hat{p}(\bx;\theta,\phi); \sigma_D\Big) \\
&=& d_{CW}^2\Bigg(\int_{\mbR^d} q(\bx|\bz;\phi) q(\bz;\phi) d\bz, \nonumber\\
&& \quad\quad\quad \int_{\mbR^d} \prod_{j=1}^D q_j(\bx_j|\bz;\phi) q(\bz;\phi) d\bz; \sigma_D\Bigg). \nonumber
\eea
We expect that this regularization will help reduce the discrepancy in \eqref{eq:ineq} that arises when Assumption \ref{assump:indep} is violated.

The incorporation of the Cramer-Wold distance regularization term of \eqref{eq:cwreg} results in a new generative model, DistVAE (+CW), and we fit DistVAE (+CW) using the following objective function:
\bean
&& \cL_{Dist+CW}(\theta,\phi) \\
&\coloneqq& \mbE_{p(\bx)} \mbE_{q(\bz|\bx;\phi)} \left[ \sum_{j=1}^D \int_0^1 \rho_{\alpha}\Big( \bx_j - Q_j(\alpha,\bz; \theta_j) \Big) d\alpha \right] \nonumber\\
&+& \lambda \cdot \log d_{CW}^2\Big(p(\bx), \hat{p}(\bx;\theta,\phi); \sigma_D\Big) \\
&+& \beta \cdot \mbE_{p(\bx)} [\mathcal{KL}(q(\bz|\bx;\phi) \| p(\bz))].
\eean

Table \ref{tab:motiv} indicates that DistVAE and CW2 exhibit comparable performance in terms of joint metrics. However, a significant disparity arises in the context of marginal distributional similarity metric performance. This is because the objective of DistVAE, denoted as $\mathcal{L}_{Dist}$, is mainly focused on learning marginal conditional distributions using the CRPS loss for each variable (element-wise reconstruction loss). 

Also, Table \ref{tab:motiv} indicates that the incorporation of Cramer-Wold distance regularization into DistVAE does result in a marginal metric performance decrease, albeit slightly. However, a notable enhancement is observed in the overall performance when considering joint metrics. Remarkably, DistVAE (+CW), despite utilizing Cramer-Wold distance as the reconstruction loss just like CW2, exhibits significantly superior overall performance. 

\textit{This underscores the significance of the distinction in the reconstruction loss between DistVAE (+CW) and CW2, the presence or absence of marginal reconstruction loss, on synthetic data generation performance}\footnote{It's worth noting that while there are differences in the regularization applied to the latent variable, with Cramer-Wold distance and KL-divergence being used, the impact is expected to be minimal due to the small latent dimension size of 2 is considered in this paper.}. The experimental results of Table \ref{tab:motiv} will be elaborated in Table \ref{tab:metrics} of Section \ref{sec:4.3}, along with explanations of the metrics and a detailed comparison. 

\textbf{Latent space.} Additionally, we visually assess the performance difference in terms of PCD between DistVAE and DistVAE (+CW) through Figure \ref{fig:latent1}. PCD evaluates the ability of the synthesizer to preserve the linear correlation structure in the generated data. Since both models generate synthetic data using inverse transform sampling, synthetic data is created by using only the median value given latent variables. The color of each point on the scatter plot represents the values of the generated $\bx_1$, while the size of each point indicates the values of the generated $\bx_2$. For (a)-(b), we use the \texttt{credit} dataset, where $\bx_1$ represents \texttt{AMT ANNUITY}, and $\bx_2$ represents \texttt{AMT GOODS PRICE}. For (c)-(d), we utilize the \texttt{loan} dataset, where $\bx_1$ corresponds to \texttt{Age}, and $\bx_2$ corresponds to \texttt{Experience}.

In Figure \ref{fig:latent1}, it is evident that the color and size of points in (b) and (d) indicate a stronger linear relationship compared to (a) and (c). Also, the PCD metric score decreases from 1.481 (DistVAE) to 0.972 (DistVAE (+CW)). This observation suggests that DistVAE (+CW) is capable of generating synthetic data while preserving the linear correlation structure in a 2-dimensional latent space, and using the Cramer-Wold distance as the reconstruction loss helps maintain statistical similarity in terms of joint distributional aspects.

\subsection{Mixture Cramer-Wold Distance}
\label{sec:3.2}

As discussed in the preceding section, it is apparent that acquiring knowledge regarding marginal distributions plays a significant role in enhancing the effectiveness of synthetic data generation. However, when utilizing the Cramer-Wold distance-based reconstruction loss, a challenge arises in the context of marginal distributional learning. This challenge stems from the fact that the point masses represented by $e_j$ on the standard basis vectors in $\mathbb{R}^D$ under the normalized surface measure $\sigma_D$ are zero. Consequently, this implies that the element-wise squared distances (i.e., the element-wise reconstruction loss) between the $j$th variable of the observation and the synthetic sample are not taken into consideration.

% The Cramer-Wold distance calculates the squared distance between two sliced samples, $\nu^\top \bx$ and $\nu^\top \hat{\bx}$, by taking into account the mean squared $L_2$ distance.
% As a result, the existing reconstruction loss of CW2 denoted as $d_{CW}^2\Big(p(\bx), \hat{p}(\bx;\theta,\phi); \sigma_D\Big)$, poses a challenge when it comes to performing marginal distributional learning. 

To effectively incorporate marginal distributional learning into the Cramer-Wold distance, we modify the integral measure of the Cramer-Wold distance by introducing the mixture measure. Denote the point mass at $e_j$ as $\delta_{e_j}$ where $e_j$ are the standard basis vectors in $\mbR^D$. Then, the mixture measure $\Tilde{\sigma}_D$ is defined as
\bean
\Tilde{\sigma}_D = \pi \cdot \left(\sum_{j=1}^D \alpha_j \delta_{e_j}\right) + (1 - \pi) \cdot \sigma_D,
\eean
where $\pi \in (0, 1)$, $\alpha_j \in (0, 1)$ for all $j$, $\pi \alpha_j, j=1,\cdots,D$ are the weights for the point masses $\delta_{e_j}$ such that $\sum_{j=1}^D \alpha_j = 1$, and $1 - \pi$ is the weight for the normalized surface measure $\sigma_D$. 

\begin{definition}[Mixture Cramer-Wold distance] \label{def:mixcw}
Let two PDFs $p(\bx)$ and $\hat{p}(\bx)$ are given, where $\bx \in \mbR^D$. The mixture Cramer-Wold distance $d_{mixCW}^2$ with $\Tilde{\sigma}_D$ is defined as
\bean
&& d_{mixCW}^2\Big(p(\bx), \hat{p}(\bx); \Tilde{\sigma}_D\Big) \\
&\coloneqq& \int_{S_D} \left\| \mbox{sm}_\gamma(\nu^\top\bX) - \mbox{sm}_\gamma(\nu^\top\hat{\bX}) \right\|_2^2 d\Tilde{\sigma}_D(\nu)\\
&=& \pi \cdot \sum_{j=1}^D \alpha_j \left\| \mbox{sm}_\gamma(e_j^\top\bX) - \mbox{sm}_\gamma(e_j^\top\hat{\bX}) \right\|_2^2 \\
&+& (1-\pi) \cdot \int_{S_D} \left\| \mbox{sm}_\gamma(\nu^\top\bX) - \mbox{sm}_\gamma(\nu^\top\hat{\bX}) \right\|_2^2 d\sigma_D(\nu),
\eean
where $\nu \in S_D$, $\nu^\top\bX \coloneqq \{\nu^\top\bx^{(i)}\}_{i=1}^n$ for $\bx^{(i)} \sim p(\bx)$, and $\nu^\top\hat{\bX} \coloneqq \{\nu^\top\hat{\bx}^{(i)}\}_{i=1}^n$ for $\hat{\bx}^{(i)} \sim \hat{p}(\bx)$.
% $\bX_{\eta_1} \coloneqq \{\bx^{(i)}\}_{i=1}^n$, $\bx^{(i)} \sim p(\bx;\eta_1)$, $\bX_{\eta_2} \coloneqq \{\bx^{(i)}\}_{i=1}^n$, $\bx^{(i)} \sim p(\bx;\eta_2)$, 
\end{definition}

Since $e_j^\top\bX = \{e_j^\top\bx^{(i)}\}_{i=1}^n = \{\bx_j^{(i)}\}_{i=1}^n$, the first term of the mixture Cramer-Wold distance in Definition \ref{def:mixcw} is re-written as
\bean
&& \sum_{j=1}^D \alpha_j \left\| \mbox{sm}_\gamma(e_j^\top\bX) - \mbox{sm}_\gamma(e_j^\top\hat{\bX}) \right\|_2^2 \\
&=& \sum_{j=1}^D \alpha_j \left\| \frac{1}{n} \sum_{i=1}^n N(\bx_j^{(i)}, \gamma) - \frac{1}{n} \sum_{i=1}^n N(\hat{\bx}_j^{(i)}, \gamma) \right\|_2^2 \\
&=& \sum_{j=1}^D \alpha_j \Bigg\{ \frac{1}{n^2} \sum_{l=1}^n \sum_{k=1}^n \frac{1}{\sqrt{4\pi\gamma}} \exp \left( -\frac{1}{4\gamma} (\bx_j^{(l)} - \bx_j^{(k)})^2 \right) \\
&+& \frac{1}{n^2} \sum_{l=1}^n \sum_{k=1}^n \frac{1}{\sqrt{4\pi\gamma}} \exp \left( -\frac{1}{4\gamma} (\hat{\bx}_j^{(l)} - \hat{\bx}_j^{(k)})^2 \right) \\
&-& \frac{2}{n^2} \sum_{l=1}^n \sum_{k=1}^n \frac{1}{\sqrt{4\pi\gamma}} \exp \left( -\frac{1}{4\gamma} (\bx^{(l)}_j - \hat{\bx}_j^{(k)})^2 \right) \Bigg\},
\eean
and it corresponds to the sum of the element-wise Cramer-Wold distance. 
 
\subsection{Cramer-Wold Distributional AutoEncoder}
\label{sec:3.3}

Based on the mixture Cramer-Wold distance (Definition \ref{def:mixcw}), we introduce a new generative model learning method. We refer to this model as CWDAE (Cramer-Wold distance Distributional AutoEncoder), and it is based on the modeling technique of DistVAE. The objective of CWDAE, denoted as $\mathcal{L}(\theta,\phi)$, is to minimize the following with respect to $\theta, \phi$:
\bea \label{eq:logCWDAEpi}
\cL(\theta,\phi) &\coloneqq& \log d_{mixCW}^2\Big(p(\bx), \hat{p}(\bx;\theta,\phi); \Tilde{\sigma}_D\Big) \nonumber\\
&+& \lambda \cdot \log d_{CW}^2\Big(p(\bz), \hat{q}(\bz;\phi); \sigma_d\Big),
% \log d_{mixCW}^2(\bX, \hat{\bX}; \Tilde{\sigma}_D) + \lambda \cdot \log d_{CW}^2(\bZ, \hat{\bZ}; \sigma_d)
\eea
where we set $\alpha_j = 1/D$ for all $j$, and treat $\pi$ of $\Tilde{\sigma}_D$ as a hyper-parameter. The synthetic data generation procedure of CWDAE is shown in Algorithm \ref{alg:syndata}

It's worth noting that \eqref{eq:logCWDAEpi} can be computed in a closed form, and the closed-form solution can be found in Appendix \ref{app:1.1}. Because \cite{Tabor2018CramerWoldA} demonstrated the benefits of applying a logarithmic function to the Cramer-Wold distance, in contrast to CW2, we applied a logarithm function not only to the Cramer-Wold distance regularization of the latent variable but also to the Cramer-Wold distance for the reconstruction loss. 

\textbf{Modeling.}
The encoder and decoder of CWDAE, when dealing with observations that comprise both continuous and discrete variables, are assumed to be quite similar to DistVAE. They are assumed to be:
\bea \label{eq:cwdae_decoder}
q(\bz|\bx;\phi) &=& \mathcal{N}\big(\bz | \mu(\bx;\phi), diag(\sigma^2(\bx;\phi))\big) \nonumber\\
p(\bx|\bz;\theta) &=& \prod_{j \in I_c} \frac{d}{d\bx_j} Q^{-1}_j(\bx_j|\bz;\theta_j) \nonumber\\
&& \times \prod_{j \in I_d} \prod_{l=1}^{T_j} \pi_l(\bz;\theta_j)^{\mathbb{I}(\bx_{jl} = 1)},
\eea
where $\pi(\cdot; \theta_j): \mbR^d \mapsto \Delta^{T_j-1}$ is a neural network parameterized with $\theta_j$ for $j \in I_d$ and the subscript $l$ refers to the $l$th element of the vector.

In the forward step, we employ the Gumbel-Softmax sampling \cite{maddison2017the, jang2017categorical} and the Straight-Through (ST) Estimator \cite{Bengio2013EstimatingOP} with the probability vector $\pi(\bz;\theta_j)$ for $j \in I_d$\footnote{In this paper, both CW2 and DistVAE-based synthesizers also utilized Gumbel-Softmax sampling and the Straight-Through (ST) Estimator in the forward step to sample discrete variables.}. The temperature of Gumbel-Softmax sampling is annealed according to the following schedule:
\bean
\max\big( 10 \cdot \exp(-0.025 \cdot epoch), \tau \big),
% tau = np.maximum(10 * np.exp(-0.025 * epoch), config["tau"])
\eean
where $\tau$ is a hyper-parameter and $epoch$ is the global training iteration.

\begin{algorithm}
\caption{Synthetic data generation procedure} 
\hspace*{\algorithmicindent} \textbf{Input} $n$: the number of synthetic samples \\
\hspace*{\algorithmicindent} \textbf{Output} $\{\Tilde{x}^{(i)}\}_{i=1}^n$: synthetic dataset 
\begin{algorithmic}[1]
    \For {$i=1,2,\ldots,n$}
        \State Sample $\bz \sim p(\bz)$
        \For {$j \in I_c$}
            \State Sample $\alpha \sim U(0, 1)$
            \State $\Tilde{x}_j^{(i)} \leftarrow Q_j(\alpha|\bz;\theta_j)$
        \EndFor
        \For {$j \in I_d$}
            \State Sample $G_l \sim_{i.i.d.} Gumbel(0, 1)$ for $l=1,\cdots,T_j$
            \State $\Tilde{x}_j^{(i)} = \arg\max_{l=1,\cdots,T_j} \{\log \pi_l(z;\theta_j) + G_l\}$
        \EndFor
        \State $\Tilde{x}^{(i)} \leftarrow (\Tilde{x}_j^{(i)})_{j \in I_c \cup I_d}$
    \EndFor
\end{algorithmic} 
\label{alg:syndata}
\end{algorithm}

\textbf{Controllable privacy level \cite{Park2018DataSB}.}
\cite{an2023distributional} effectively handles the trade-off between the quality of synthetic data and the risk of privacy leakage by controlling the hyper-parameter $\beta$, which is the hyper-parameter determines the weight of the KL-divergence term within the ELBO. Similarly, in our proposed method, CWDAE, we can also strike a balance between the quality of synthetic data, considering both marginal and joint distributional aspects, and the potential risks of privacy leakage. This balance can be achieved by adjusting the hyper-parameter $\pi$, which controls the weight of the marginal reconstruction loss within the mixture Cramer-Wold distance. The practical demonstration of this capability is presented in the experimental results in Table \ref{tab:privacy} in Section \ref{sec:4.3}.

\subsubsection{Comparison to Prior Work}
The difference in the parameterization of the generative model between CW2 and CWDAE significantly impacts the distributional capacity that each model can represent. The generative model derived from the CW2 model, which is based on the AutoEncoder, can be written as:
\bea \label{eq:cw2gen}
\int_{\mbR^d} \delta(\bx - G(\bz;\theta)) p(\bz) d\bz.
\eea 
On the other hand, for continuous variables, the generative model of CWDAE (also DistVAE) is written as follows:
\bea \label{eq:distgen}
\int_{\mbR^d} \prod_{j=1}^D \frac{d}{d\bx_j} Q^{-1}_j(\bx_j|\bz;\theta_j) p(\bz) d\bz.
\eea
Comparing \eqref{eq:cw2gen} and \eqref{eq:distgen}, since the decoder in \eqref{eq:cw2gen} is the Dirac delta function and the decoder in \eqref{eq:distgen} is the arbitrary density function $\frac{d}{d\bx_j} Q^{-1}_j(\bx_j|\bz;\theta_j)$, \eqref{eq:distgen} can represent a broader range of distributions compared to \eqref{eq:cw2gen}. 

This distinction becomes more apparent during the synthetic data sampling process. In the case of CW2, when the latent variable $\bz$ is sampled from the prior distribution, the synthetic data is fixed to a single value, determined by $G(\bz;\theta)$. However, in the case of CWDAE, even when $\bz$ is fixed, the inverse transform sampling method varies the synthetic sample based on the value of $\alpha \in (0, 1)$ (i.e., quantile level). This results in more diverse synthetic samples for a given $\bz$. The difference in synthetic data sample diversity is empirically demonstrated in Table \ref{tab:privacy} of Section \ref{sec:4.3}.

\section{Experiments}
\label{sec:4}

We run all experiments using Geforce RTX 3090 GPU, and our experimental codes are all available with \texttt{pytorch}\footnote{We release the code at \url{https://github.com/an-seunghwan/XXX}.}. 

\subsection{Overview}
\label{sec:4.1}

\begin{table}[h]
\caption{Description of datasets. $|I_c|$ represents the number of continuous and ordinal variables. $|I_d|$ denotes the number of discrete variables. $D$ is the dimension of the observation.}
  \centering
  \begin{tabular}{lrrrrrr}
    \toprule
    Dataset & Train/Test Split & $|I_c|$ & $|I_d|$ & $D$ \\
    \midrule
\texttt{covtype} & 45k/5k & 10 & 1 & 17\\
\texttt{credit} & 45k/5k & 10 & 9 & 49\\
\texttt{loan} & 4k/1k & 5 & 6 & 19\\
\texttt{adult} & 40k/5k & 5 & 9 & 105\\
\texttt{cabs} & 40k/1k & 6 & 7 & 52\\
\texttt{kings} & 20k/1k & 11 & 7 & 84\\
    \bottomrule
  \end{tabular}
\label{tab:data_description}
\end{table}

\textbf{Datasets.} 
For evaluation, we consider the following six readily available real tabular datasets: \texttt{covtype}, \texttt{credit}, \texttt{loan}, \texttt{adult}, \texttt{cabs}, and \texttt{kings} (URL links for these datasets can be found in Table \ref{tab:urls} Appendix \ref{app:1}). The detailed data descriptions are in Table \ref{tab:data_description}. 
The datasets \texttt{covtype} and \texttt{loan} have a dimensionality of $D < 20$. In these cases, the closed-form calculation of the Cramer-Wold distance, which relies on Kummer's confluent hypergeometric function's asymptotic formula \cite{Tabor2018CramerWoldA}, may not be applicable. Nevertheless, in this paper, we include these datasets in our experiments since we confirm that our proposed model CWDAE still yields good performance empirically, even when using the asymptotic formula.
% We treat the ordinal variables as continuous variables and discretize the estimated CDF (see Appendix \ref{app:5} for the discretization algorithm). 

\textbf{Compared models.}
We compare CWDAE with the state-of-the-art synthesizers: CTGAN \cite{xu2019ctgan}, TVAE \cite{xu2019ctgan}, CTAB-GAN \cite{zhao2021ctabgan}, CW2 \cite{Knop2020GenerativeMW}\footnote{While \cite{Knop2020GenerativeMW} proposed LCW, which is based on CW2 and employs a two-stage learning approach, we focused on comparing only one-stage models in this paper.}, and DistVAE \cite{an2023distributional}. 
The selected latent space dimension ($d=2$) is the same for all models. This choice intentionally restricts the decoder's capacity in all models. However, we maintain that the number of parameters remains small and consistent across all models throughout the experiment. This approach allows us to isolate the performance differences in synthetic data generation and emphasize the impact of each synthesizer's methodology, particularly highlighting the contribution of the discrepancy measure for the reconstruction loss. For more detailed model architecture and a comprehensive comparison of the number of model parameters, please refer to Table \ref{tab:arch} and Table \ref{tab:num_params} in Appendix \ref{app:1}.

\begin{table*}[ht]
\caption{Hyper-parameter settings.}
\centering
\resizebox{0.8\linewidth}{!}{
  \begin{tabular}{lrrrrrrrr}
    \toprule
    Model & epochs & batch size & learning rate & $\beta$ (or decoder std range) & $\lambda$ & $\tau$ & $d$ \\
    \midrule
CTGAN & 300 & 500 & 0.0002 & - & - & - & 2  \\
TVAE & 200 & 256 & 0.005 & [0.1, 1] & - & - & - 2 \\
CTAB-GAN & 150 & 500 & 0.0002 & - & - & - & 2 \\
CW2 & 100 & 1024 & 0.001 & - & 1 & 0.2 & 2 \\
DistVAE & 100 & 256 & 0.001 & 0.5 & - & 0.2 & 2  \\
\midrule
CW2 (+Q) & 100 & 1024 & 0.001 & - & 1 & 0.2 & 2 \\
CW2 (+E) & 100 & 1024 & 0.001 & - & 1 & 0.2 & 2  \\
CW2 (+E,Q) & 100 & 1024 & 0.001 & - & 1 & 0.2 & 2  \\
CW2 (log, +E,Q) & 100 & 1024 & 0.001 & - & 1 & 0.2 & 2  \\
DistVAE (+CW) & 100 & 1024 & 0.001 & 0.5 & 1 & 0.2 & 2  \\
\midrule
CWDAE ($\pi=0.05$) & 100 & 1024 & 0.001 & - & 1 & 0.2 & 2  \\
CWDAE ($\pi=0.1$) & 100 & 1024 & 0.001 & - & 1 & 0.2 & 2  \\
CWDAE ($\pi=0.5$) & 100 & 1024 & 0.001 & - & 1 & 0.2 & 2  \\
CWDAE ($\pi=0.9$) & 100 & 1024 & 0.001 & - & 1 & 0.2 & 2  \\
    \bottomrule
  \end{tabular}
}
\label{tab:hyperparams}
\end{table*}

\textbf{Hyper-parameters.}
For CWDAE, we conducted a grid search over batch sizes in \{512, 1024\} and learning rates in \{0.001, 0.005\}. To assess the impact of $\pi$ on CWDAE's synthetic data generation performance, we trained the model with various values of $\pi$ from \{0.05, 0.1, 0.5, 0.9\}. The weight parameter $\lambda$ was consistently set to 1 for all tabular datasets without specific tuning for each dataset. We chose this experimental setup based on the findings of \cite{Tabor2018CramerWoldA}, which demonstrated that $\lambda=1$ is sufficient for achieving satisfactory performance across most scenarios. This choice also highlights the robustness of our proposed CWDAE model in achieving consistent synthetic data generation performance across diverse datasets, regardless of the specific value of the hyper-parameter $\lambda$. The hyper-parameter choices of all models can be found in Table \ref{tab:hyperparams}.

\begin{table}[t]
\caption{Summary of the models for the ablation study.}
  \centering
  \resizebox{0.9\columnwidth}{!}{
  \begin{tabular}{lccccccc}
    \toprule
    Model & E & Q & CW & log & mix\\
    \midrule
    CW2 & \color{red}{\ding{55}} & \color{red}{\ding{55}} & \color{blue}{\ding{51}} & \color{red}{\ding{55}} & \color{red}{\ding{55}} \\
    DistVAE & \color{blue}{\ding{51}} & \color{blue}{\ding{51}} & \color{red}{\ding{55}} & \color{red}{\ding{55}} & \color{red}{\ding{55}} \\
    \midrule
    CW2 (+Q) & \color{red}{\ding{55}} & \color{blue}{\ding{51}} & \color{blue}{\ding{51}} & \color{red}{\ding{55}} & \color{red}{\ding{55}} \\
    CW2 (+E) & \color{blue}{\ding{51}} & \color{red}{\ding{55}} & \color{blue}{\ding{51}} & \color{red}{\ding{55}} & \color{red}{\ding{55}} \\
    CW2 (+E,Q) & \color{blue}{\ding{51}} & \color{blue}{\ding{51}} & \color{blue}{\ding{51}} & \color{red}{\ding{55}} & \color{red}{\ding{55}} \\
    CW2 (log, +E,Q) & \color{blue}{\ding{51}} & \color{blue}{\ding{51}} & \color{blue}{\ding{51}} & \color{blue}{\ding{51}} & \color{red}{\ding{55}} \\
    DistVAE (+CW) & \color{blue}{\ding{51}} & \color{blue}{\ding{51}} & \color{blue}{\ding{51}} & \color{blue}{\ding{51}} & \color{red}{\ding{55}} \\
    \midrule
    CWDAE ($\pi=0.05$) & \color{blue}{\ding{51}} & \color{blue}{\ding{51}} & \color{blue}{\ding{51}} & \color{blue}{\ding{51}} & \color{blue}{\ding{51}} \\
    CWDAE ($\pi=0.1$) & \color{blue}{\ding{51}} & \color{blue}{\ding{51}} & \color{blue}{\ding{51}} & \color{blue}{\ding{51}} & \color{blue}{\ding{51}} \\
    CWDAE ($\pi=0.5$) & \color{blue}{\ding{51}} & \color{blue}{\ding{51}} & \color{blue}{\ding{51}} & \color{blue}{\ding{51}} & \color{blue}{\ding{51}} \\
    CWDAE ($\pi=0.9$) & \color{blue}{\ding{51}} & \color{blue}{\ding{51}} & \color{blue}{\ding{51}} & \color{blue}{\ding{51}} & \color{blue}{\ding{51}} \\
    \bottomrule
  \end{tabular}}
\label{tab:ablation}
\end{table}

\textbf{Ablation study.}
In addition to the models mentioned above, we conducted an ablation study by constructing additional model variants to investigate the impact of various model design choices on the synthetic data generation performance. These design choices include the presence or absence of marginal reconstruction loss and using quantile functions for decoder modeling, among others. The configurations of the models used for the ablation study are in Table \ref{tab:ablation}. 

In Table \ref{tab:ablation}, the `+E' notation indicates that we use a multivariate Gaussian distribution for the encoder (posterior distribution) instead of the deterministic encoder. The `+Q' notation signifies that we constructed the decoder using quantile functions instead of the deterministic decoder. Additionally, `CW' indicates using the Cramer-Wold distance as the reconstruction loss, while `log' denotes whether a logarithm function was applied to the Cramer-Wold reconstruction loss. Finally, `mix' indicates whether a mixture measure, as described in Definition \ref{def:mixcw}, was utilized as the integral measure for the Cramer-Wold reconstruction loss.

% \begin{table}[t]
% \caption{Modeling differences between DistVAE and CW2.}
%   \centering
%   \resizebox{\columnwidth}{!}{
%   \begin{tabular}{lcccc}
%     \toprule
%      & encoder & decoder \\
%     \midrule
% -, - & $\delta(\bz - \mu(\bx;\phi))$ & $\delta(\bx - G(\bz;\theta))$ \\
% +E, +Q & $\mathcal{N}\big(\bz | \mu(\bx;\phi), diag(\sigma^2(\bx;\phi))\big)$ & $\prod_{j=1}^D Q_j(\alpha|\bz;\theta_j)$ \\
%     \bottomrule
%   \end{tabular}
%   }
% \label{tab:modeling}
% \end{table}

\begin{table}[ht]
\caption{The configurations of regressor and classifier used to evaluate the machine learning utility. The names of all parameters used in the description are consistent with those defined in corresponding packages.}
  \centering
  \resizebox{\columnwidth}{!}{
  \begin{tabular}{ccc}
    \toprule
    Task & Configuration \\
    \midrule
    Regression & \makecell{\texttt{sklearn.ensemble.RandomForestRegressor}, \\ random\_state=0, defaulted values} \\
    \midrule
    Classification & \makecell{\texttt{sklearn.ensemble.RandomForestClassifier}, \\ random\_state=0, defaulted values} \\
    \bottomrule
  \end{tabular}}
\label{tab:mlu_setting}
\end{table}

\subsection{Evaluation Metrics}
\label{sec:4.2}

To assess the quality of the synthetic data, we employ two types of assessment criteria: 1) statistical similarity and 2) privacy preservability. Each criterion is evaluated using multiple metrics, and the performance of synthesizers is reported by averaged metrics over the six real tabular datasets. The synthetic dataset is generated to have an equal number of samples as the real training dataset. Synthetic samples of ordinal variables are rounded to the first decimal place. 

\subsubsection{Statistical Similarity}

\textbf{Marginal.}
The marginal distributional similarity between the real training and synthetic datasets is evaluated using two metrics: the two-sample Kolmogorov-Smirnov test statistic and the 1-Wasserstein distance. 
% These metrics measure the distance between the empirical marginal CDFs. 

The \textit{two-sample Kolmogorov-Smirnov test statistic} and the \textit{1-Wasserstein distance} are computed independently for each variable, measuring the distance between the real training and synthetic empirical marginal CDFs. These metrics quantify the discrepancy between the two CDFs, with both being zero when the distributions are identical and larger values indicating greater dissimilarity.

\textbf{Joint.}
The joint distributional similarity between the real training and synthetic datasets is evaluated using these four metrics: the pairwise correlation difference \cite{Goncalves2020GenerationAE, zhao2021ctabgan, an2023distributional}, log-cluster \cite{Goncalves2020GenerationAE}, mean absolute percentage error for regression task \cite{Park2018DataSB}, and the $F_1$ score for classification tasks \cite{xu2019ctgan, zhao2021ctabgan, Park2018DataSB, Choi2017GeneratingMD, Fang2022OvercomingCO}. 

The correlation coefficient is employed to evaluate the level of linear correlation captured among the variables. The \textit{pairwise correlation difference} (PCD) quantifies the difference in terms of the Frobenius norm between these correlation matrices calculated between the real training and synthetic datasets. A smaller PCD score indicates that the synthetic data closely approximates the real data in terms of linear correlations among the variables. In essence, it assesses how effectively the method captures the linear relationships between variables present in the real dataset.

The \textit{log-cluster} metric evaluates how similar the underlying structure between the real training and synthetic datasets is, with particular attention to clustering patterns. To calculate this metric, we initially combine the real training and synthetic datasets into a unified dataset. Subsequently, we apply cluster analysis to this merged dataset using the $K$-means algorithm, using a predefined number of clusters denoted as $G$. The metric is computed as follows:
\bean
\log \left( \frac{1}{G} \sum_{i=1}^G \left( \frac{n_i^R}{n_i} - c \right)^2 \right),
\eean
where $n_i$ is the number of samples in the $i$th cluster, $n^R_i$ is the number of samples from the real dataset in the $i$th cluster, and $c = n^R / (n^R + n^S)$. $n^R$ and $n^S$ denote the number of samples from the real training and synthetic dataset. In this paper, $c$ is set to 0.5 because we have $n^R = n^S$. Large values of the log-cluster metric indicate discrepancies in cluster memberships, suggesting differences in real and synthetic data distribution. As in \cite{Goncalves2020GenerationAE}, the number of clusters is set to 20.

To assess how effectively a synthetic dataset replicates the non-linear dependence structures found in the real training dataset, we utilize the \textit{mean absolute percentage error} (MAPE) and the \textit{$F_1$ score}. It can be seen as the machine learning utility, which is measured by the predictive performance of the trained model over the synthetic data. For each variable, we train a Random Forest that performs regression or classification using all variables except the one currently under consideration (one-vs-all). Subsequently, we assess the predictive performance of the excluded variable on a test dataset. Finally, we average the predictive performance metric results for all variables. The configuration of the regressor and classifier used to evaluate synthetic data quality can be found in Table \ref{tab:mlu_setting}.

\begin{table*}[t]
\caption{Marginal and joint distributional similarity. $\uparrow$ denotes higher is better and $\downarrow$ denotes lower is better.}
  \centering
  \resizebox{\textwidth}{!}{
  \begin{tabular}{lrrrrrrrrrrrrrrrr}
    \toprule
    & \multicolumn{2}{c}{marginal} & \multicolumn{4}{c}{joint} \\
    \cmidrule(lr){2-3} \cmidrule(lr){4-7}
    Model & KS $\downarrow$ & W1 $\downarrow$ & PCD $\downarrow$ & log-cluster $\downarrow$ & MAPE $\downarrow$ & $F_1$ $\uparrow$ & rank \\
    \midrule
CTAB-GAN & $0.309_{\pm 0.193}(14)$ & $1.416_{\pm 1.360}(14)$ & $2.573_{\pm 0.515}(13)$ & $-2.995_{\pm 0.406}(13)$ & $1.241_{\pm 2.271}(14)$ & $0.249_{\pm 0.279}(14)$ & 13.7\\
CTGAN & $0.104_{\pm 0.039}(12)$ & $0.162_{\pm 0.065}(12)$ & $2.179_{\pm 0.468}(12)$ & $-3.704_{\pm 0.920}(11)$ & $0.662_{\pm 1.239}(13)$ & $0.541_{\pm 0.155}(13)$ & 12.2\\
TVAE & $0.192_{\pm 0.000}(13)$ & $0.232_{\pm 0.000}(13)$ & $3.789_{\pm 0.000}(14)$ & $-2.638_{\pm 0.000}(14)$ & $0.241_{\pm 0.000}(7)$ & $0.760_{\pm 0.000}(1)$ & 10.3\\
CW2 & $0.081_{\pm 0.029}(11)$ & $0.093_{\pm 0.025}(11)$ & $1.348_{\pm 0.810}(9)$ & $-3.983_{\pm 1.334}(10)$ & $0.318_{\pm 0.344}(11)$ & $0.644_{\pm 0.085}(4)$ & 9.3\\
DistVAE & $0.056_{\pm 0.025}(4)$ & $0.066_{\pm 0.023}(5)$ & $1.481_{\pm 0.386}(10)$ & $-3.642_{\pm 0.751}(12)$ & $0.301_{\pm 0.194}(10)$ & $0.670_{\pm 0.074}(2)$ & 7.2\\
\midrule
CW2 (+Q) & $0.067_{\pm 0.030}(9)$ & $0.085_{\pm 0.039}(8)$ & $1.009_{\pm 0.367}(4)$ & $-4.830_{\pm 1.723}(5)$ & $0.242_{\pm 0.108}(8)$ & $0.639_{\pm 0.099}(7)$ & 6.8\\
CW2 (+E) & $0.078_{\pm 0.030}(10)$ & $0.093_{\pm 0.031}(10)$ & $1.263_{\pm 0.721}(8)$ & $-4.182_{\pm 1.495}(8)$ & $0.238_{\pm 0.107}(4)$ & $0.639_{\pm 0.087}(8)$ & 8.0\\
CW2 (+E,Q) & $0.066_{\pm 0.029}(8)$ & $0.086_{\pm 0.038}(9)$ & $1.018_{\pm 0.377}(5)$ & $-4.754_{\pm 1.713}(7)$ & $0.241_{\pm 0.112}(6)$ & $0.640_{\pm 0.098}(6)$ & 6.8\\
CW2 (log, +E,Q) & $0.065_{\pm 0.029}(7)$ & $0.075_{\pm 0.031}(7)$ & $0.926_{\pm 0.354}(1)$ & $-5.050_{\pm 1.748}(4)$ & $0.226_{\pm 0.091}(1)$ & $0.638_{\pm 0.097}(10)$ & 5.0\\
DistVAE (+CW) & $0.059_{\pm 0.022}(6)$ & $0.071_{\pm 0.028}(6)$ & $0.972_{\pm 0.331}(3)$ & $-4.828_{\pm 1.349}(6)$ & $0.232_{\pm 0.112}(2)$ & $0.658_{\pm 0.080}(3)$ & \underline{4.3}\\
\midrule
CWDAE ($\pi=0.05$) & $0.055_{\pm 0.026}(3)$ & $0.057_{\pm 0.021}(4)$ & $0.966_{\pm 0.364}(2)$ & $-5.212_{\pm 1.747}(1)$ & $0.234_{\pm 0.094}(3)$ & $0.640_{\pm 0.096}(5)$ & \textbf{3.0}\\
CWDAE ($\pi=0.1$) & $0.052_{\pm 0.026}(2)$ & $0.051_{\pm 0.019}(3)$ & $1.020_{\pm 0.367}(6)$ & $-5.210_{\pm 1.708}(2)$ & $0.239_{\pm 0.097}(5)$ & $0.639_{\pm 0.095}(9)$ & 4.5\\
CWDAE ($\pi=0.5$) & $0.051_{\pm 0.029}(1)$ & $0.043_{\pm 0.016}(2)$ & $1.231_{\pm 0.410}(7)$ & $-5.098_{\pm 1.607}(3)$ & $0.259_{\pm 0.111}(9)$ & $0.629_{\pm 0.102}(11)$ & 5.5\\
CWDAE ($\pi=0.9$) & $0.057_{\pm 0.038}(5)$ & $0.042_{\pm 0.018}(1)$ & $1.681_{\pm 0.392}(11)$ & $-4.124_{\pm 1.440}(9)$ & $0.324_{\pm 0.182}(12)$ & $0.620_{\pm 0.106}(12)$ & 8.3\\
    \bottomrule
  \end{tabular}
  }
\label{tab:metrics}
\end{table*}

\begin{figure*}
    \centering
    \subfigure[CWDAE ($\pi=0.05$)]{
    \includegraphics[width=0.23\textwidth]{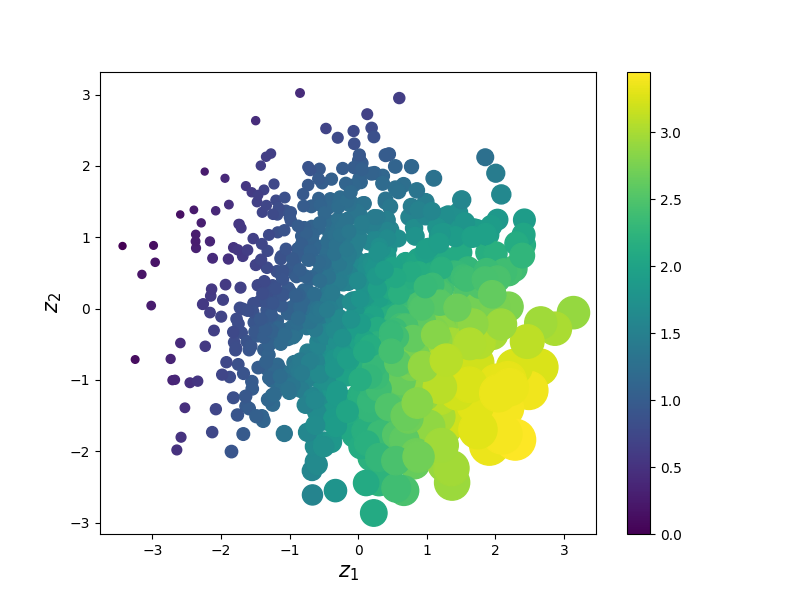}
    }
    \subfigure[CWDAE ($\pi=0.9$)]{
    \includegraphics[width=0.23\textwidth]{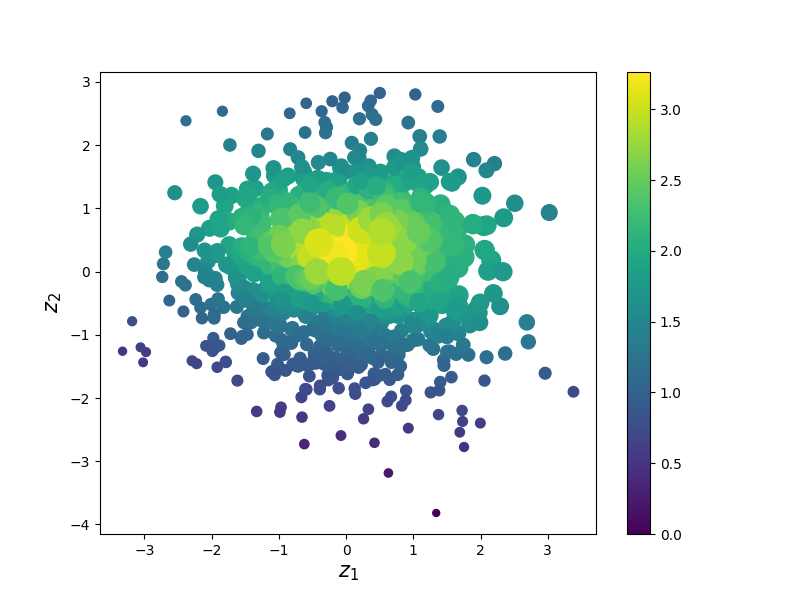}
    }
    \subfigure[CWDAE ($\pi=0.05$)]{
    \includegraphics[width=0.23\textwidth]{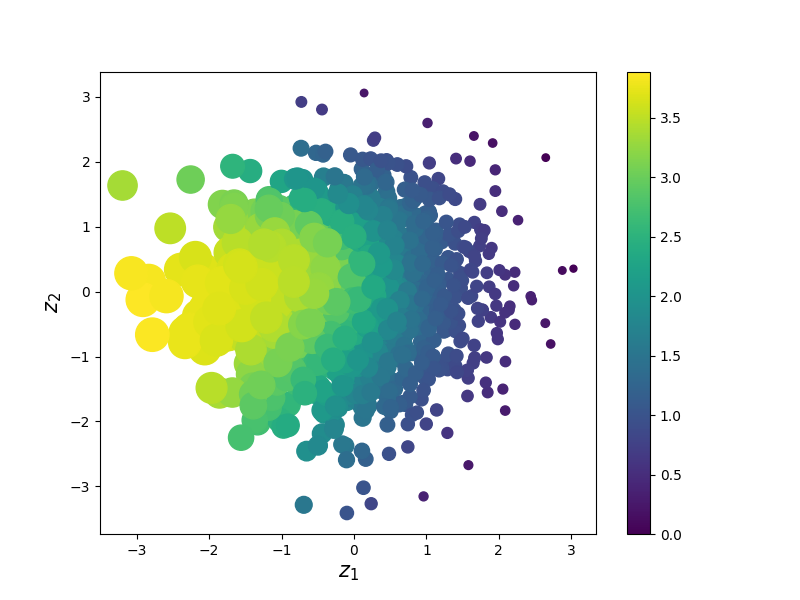}
    }
    \subfigure[CWDAE ($\pi=0.9$)]{
    \includegraphics[width=0.23\textwidth]{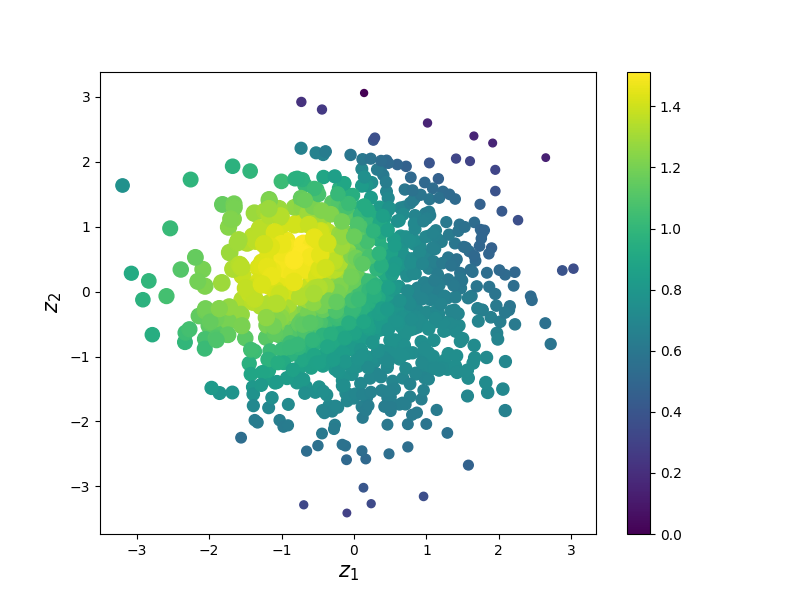}
    }
    \caption{
    Scatter plots of the latent variables and synthetic samples, which are generated from CWDAE ($\pi=0.05$) and CWDAE ($\pi=0.9$).
    % The figure visualizes the relationship between the latent space and generated samples of CWDAE ($\pi=0.05$) and CWDAE ($\pi=0.9$). In this context, synthetic data is generated by taking only the median when each latent variable is given. The values of the generated $\bx_1$ are represented by the color of each point on the scatter plot, while the size of each point indicates the values of the generated $\bx_2$. (a)-(b): \texttt{credit} dataset, $\bx_1$: AMT ANNUITY, $\bx_2$: AMT GOODS PRICE. (c)-(d): \texttt{loan} dataset, $\bx_1$: Age, $\bx_2$: Experience.
    % \textcolor{blue}{위의 3가지 데이터셋은 PCD 측면에서 특히 Cramer-Wold distance를 사용하였을 때 성능향상이 많이 이루어짐.}
    }
    \label{fig:latent2}
\end{figure*}

\begin{table*}[t]
\caption{Privacy preservability. $\uparrow$ denotes higher is better and $\downarrow$ denotes lower is better.}
  \centering
  \resizebox{\textwidth}{!}{
  \begin{tabular}{lrrrrrrrrrrrrrrrr}
    \toprule
    Model & DCR(R,S) $\uparrow$ & DCR(S,S) $\uparrow$ & AD($F_1$, 1) $\downarrow$ & AD($F_1$, 10) $\downarrow$ & AD($F_1$, 100) $\downarrow$ & rank \\
    \midrule
CTAB-GAN & $0.508_{\pm 0.258}(2)$ & $0.041_{\pm 0.075}(12)$ & $0.109_{\pm 0.115}(1)$ & $0.111_{\pm 0.117}(1)$ & $0.112_{\pm 0.119}(1)$ & \textbf{1.2}\\
CTGAN & $0.427_{\pm 0.230}(6)$ & $0.355_{\pm 0.205}(7)$ & $0.265_{\pm 0.091}(2)$ & $0.281_{\pm 0.087}(3)$ & $0.278_{\pm 0.090}(4)$ & \underline{3.8}\\
TVAE & $0.577_{\pm 0.000}(1)$ & $0.364_{\pm 0.000}(6)$ & $0.303_{\pm 0.000}(9)$ & $0.306_{\pm 0.000}(10)$ & $0.304_{\pm 0.000}(13)$ & 8.2\\
CW2 & $0.505_{\pm 0.242}(3)$ & $0.018_{\pm 0.009}(14)$ & $0.317_{\pm 0.085}(13)$ & $0.314_{\pm 0.079}(12)$ & $0.304_{\pm 0.070}(12)$ & 10.0\\
DistVAE & $0.441_{\pm 0.248}(5)$ & $0.461_{\pm 0.282}(1)$ & $0.318_{\pm 0.081}(14)$ & $0.315_{\pm 0.066}(13)$ & $0.301_{\pm 0.061}(11)$ & 10.8\\
\midrule
CW2 (+Q) & $0.415_{\pm 0.192}(8)$ & $0.383_{\pm 0.176}(4)$ & $0.303_{\pm 0.080}(10)$ & $0.302_{\pm 0.071}(9)$ & $0.292_{\pm 0.068}(9)$ & 9.0\\
CW2 (+E) & $0.501_{\pm 0.238}(4)$ & $0.019_{\pm 0.009}(13)$ & $0.310_{\pm 0.085}(11)$ & $0.312_{\pm 0.071}(11)$ & $0.304_{\pm 0.068}(14)$ & 10.0\\
CW2 (+E,Q) & $0.418_{\pm 0.192}(7)$ & $0.381_{\pm 0.170}(5)$ & $0.302_{\pm 0.082}(8)$ & $0.301_{\pm 0.074}(8)$ & $0.287_{\pm 0.072}(8)$ & 7.8\\
CW2 (log, +E,Q) & $0.390_{\pm 0.196}(11)$ & $0.348_{\pm 0.176}(9)$ & $0.301_{\pm 0.088}(7)$ & $0.301_{\pm 0.075}(7)$ & $0.285_{\pm 0.068}(5)$ & 7.5\\
DistVAE (+CW) & $0.408_{\pm 0.216}(10)$ & $0.408_{\pm 0.231}(2)$ & $0.310_{\pm 0.081}(12)$ & $0.316_{\pm 0.072}(14)$ & $0.297_{\pm 0.068}(10)$ & 11.5\\
\midrule
CWDAE ($\pi=0.05$) & $0.374_{\pm 0.200}(12)$ & $0.341_{\pm 0.197}(11)$ & $0.297_{\pm 0.079}(5)$ & $0.296_{\pm 0.073}(5)$ & $0.285_{\pm 0.061}(6)$ & 7.0\\
CWDAE ($\pi=0.1$) & $0.370_{\pm 0.201}(14)$ & $0.341_{\pm 0.202}(10)$ & $0.297_{\pm 0.085}(6)$ & $0.300_{\pm 0.068}(6)$ & $0.286_{\pm 0.061}(7)$ & 8.2\\
CWDAE ($\pi=0.5$) & $0.373_{\pm 0.211}(13)$ & $0.353_{\pm 0.209}(8)$ & $0.291_{\pm 0.080}(4)$ & $0.284_{\pm 0.072}(4)$ & $0.274_{\pm 0.066}(3)$ & 6.0\\
CWDAE ($\pi=0.9$) & $0.409_{\pm 0.228}(9)$ & $0.397_{\pm 0.231}(3)$ & $0.284_{\pm 0.096}(3)$ & $0.278_{\pm 0.081}(2)$ & $0.265_{\pm 0.076}(2)$ & 4.0\\
    \bottomrule
  \end{tabular}
  }
\label{tab:privacy}
\end{table*}

\subsubsection{Privacy preservability}

The privacy-preserving capacity is measured using two metrics: the distance to the closest record \cite{Park2018DataSB, zhao2021ctabgan, an2023distributional} and attribute disclosure \cite{Choi2017GeneratingMD, an2023distributional}.

As in \cite{zhao2021ctabgan}, the \textit{distance to closest record} (DCR) is defined as the $5^{th}$ percentile of the $L_2$ distances between all real training samples and synthetic samples. Since the DCR is a $L_2$ distance-based metric, it is computed using only continuous variables. A higher DCR value indicates a more effective preservation of privacy, indicating a lack of overlap between the real training data and the synthetic samples. Conversely, an excessively large DCR score suggests a lower quality of the generated synthetic dataset. Therefore, the DCR metric provides insights into both the privacy-preserving capability and the quality of the synthetic dataset.

\textit{Attribute disclosure} pertains to a scenario where potential attackers can unveil extra attributes of a record by leveraging a subset of attributes they already possess, along with similar records extracted from the synthetic dataset. These similar records, often referred to as nearest neighbors, are determined based on the $L_2$ distance for continuous variables. In this study, we consider varying numbers of nearest neighbors, specifically 1, 10, and 100. To quantify how accurately attackers can identify these additional attributes, we employ the $F_1$ score, which assesses the extent to which discrete variables are revealed by attackers. Higher attribute disclosure metrics signify a high risk of privacy violation, indicating that attackers can effectively deduce undisclosed variables. 
% From a privacy standpoint, attribute disclosure can be viewed as a more significant concern compared to membership inference attacks, as it assumes that attackers have access to only a subset of attributes for a given record \cite{Choi2017GeneratingMD}.

\subsection{Results}
\label{sec:4.3}

In this section, the metric scores in the result tables are based on the 10 repeated experiments conducted for each dataset. These results are then summarized by calculating the mean and standard deviation across all datasets.

For notational simplicity, we abbreviate the metrics as follows: the two-sample Kolmogorov-Smirnov test statistic as KS, the 1-Wasserstein distance as W1, the pairwise correlation difference as PCD, log-cluster as log-cluster, the mean absolute percentage error for regression task as MAPE, the $F_1$ score for classification task as $F_1$, the distance to closest record between the real training and the synthetic dataset as DCR(R,S), the distance to closest record within the synthetic dataset as DCR(S,S), and the attribute disclosure score ($F_1$ score) with 1, 10, 100 nearest neighbors as AD($F_1$, 1), AD($F_1$, 10), AD($F_1$, 100), respectively. 

Additionally, the numbers enclosed in parentheses next to the metric scores indicate the ranking of each model with respect to that specific metric. The `rank' column provides an overall ranking for each model, which is determined by averaging the ranks assigned to each model across all metrics.

\subsubsection{Statistical similarity}

Table \ref{tab:metrics} presents the results of statistical similarity metrics for all models. Firstly, when comparing CWDAE to five baseline models (CTAB-GAN, CTGAN, TVAE, CW2, and DistVAE), CWDAE ($\pi=0.05$) achieves the best average ranking of 3.0. Also, CWDAE-based synthesizers generally outperform these baseline models in terms of average ranking, except for the case with $\pi=0.9$. This indicates that the proposed CWDAE model consistently exhibits superior performance in both marginal and joint metrics.

Regarding modeling techniques, comparing CW2 to CW2 (+E), we observe an overall improvement in performance, highlighting the benefits of using a stochastic encoder. Similarly, when comparing CW2 to CW2 with quantile function modeling in the decoder (CW2 vs. CW2 (+Q)), quantile modeling leads to significant performance enhancements across most metrics. Therefore, from a modeling perspective, both quantile modeling and a stochastic encoder contribute to improved performance, with quantile modeling being particularly effective. 

Furthermore, when comparing CW2 (+E,Q) to CW2 (log, +E,Q), we observe performance improvements in most metrics for CW2 (log, +E,Q). This suggests that applying a logarithmic function to the Cramer-Wold distance used as the reconstruction loss is also beneficial for synthetic data generation performance.

CWDAE shares the same encoder and decoder modeling with CW2 (log, +E,Q), and both use the logarithmic function-applied Cramer-Wold distance as the reconstruction loss. However, CWDAE incorporates a marginal reconstruction loss by adopting the mixture Cramer-Wold distance in the reconstruction loss. Therefore, when comparing the performance of CWDAE and CW2 (log, +E,Q), the overall better performance of CWDAE (the average ranking of CW2 (log, +E,Q) (5.0) is lower than that of CWDAE ($\pi=0.05$) (3.0)) indicates that element-wise reconstruction loss contributes to improving synthetic data generation performance.

Comparing CWDAE to the top-performing DistVAE (+CW) among the nine models (the average ranking is 4.3), including baseline and ablation study models, CWDAE exhibited superior performance when $\pi=0.05$. This indicates that the Cramer-Wold distance-based marginal reconstruction loss is more effective in terms of synthetic data generation performance than DistVAE's CRPS loss-based marginal reconstruction loss.

Furthermore, it is observed that as the value of $\pi$ increases, which determines the weight of the marginal Cramer-Wold distance, there is a gradual decline in joint metric performance and a gradual enhancement in marginal metric performance. This observation suggests that $\pi$ can be effectively used to control the balance between the statistical similarity of marginal and joint perspectives, allowing for a trade-off adjustment between these aspects.
% It's worth mentioning that utilizing both of these features allows for a direct comparison between CWDAE and DistVAE-based models since they share the same model structure.

\textbf{Latent space.} 
Similarly to Figure \ref{fig:latent1} in Section \ref{sec:3.1}, we visually assess the performance difference in terms of PCD between CWDAE ($\pi=0.05$) and CWDAE ($\pi=0.9$) through Figure \ref{fig:latent2}. Since both models utilize quantile functions for decoder modeling, synthetic data is generated fairly by considering only the median value for each latent variable. The color of each point on the scatter plot represents the values of the generated $\bx_1$, while the size of each point indicates the values of the generated $\bx_2$. For (a)-(b), we use the \texttt{credit} dataset, where $\bx_1$ represents \texttt{AMT ANNUITY}, and $\bx_2$ represents \texttt{AMT GOODS PRICE}. For (c)-(d), we utilize the \texttt{loan} dataset, where $\bx_1$ corresponds to \texttt{Age}, and $\bx_2$ corresponds to \texttt{Experience}.

In Figure \ref{fig:latent2}, it is evident that the color and size of points in (a) and (c) indicate a stronger linear relationship on the latent space compared to (b) and (d). This aligns with the observation that as $\pi$ increases from 0.05 to 0.9, the PCD metric score also increases from 0.966 to 1.681, indicating a deterioration in PCD performance. In other words, this observation suggests that CWDAE ($\pi=0.05$) is capable of generating synthetic data while preserving the linear correlation structure, and $\pi$ can determine the performance of statistical similarity in terms of joint distributional aspects.

\begin{figure}[h]
    \centering
    \includegraphics[width=0.99\columnwidth]{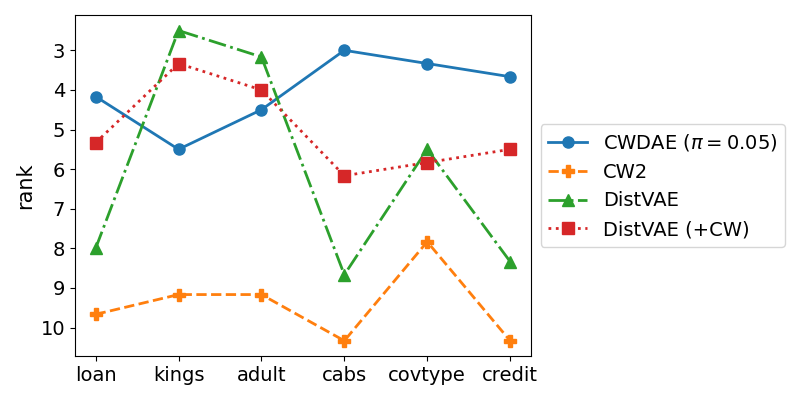}
    \caption{The average rank of distributional similarity metrics for each tabular dataset. The x-axis of the plot is sorted in ascending order based on the size of the training dataset.}
    \label{fig:trainingsetsize}
\end{figure}

\textbf{Effect of Training Dataset Size.}
Additionally, in Figure \ref{fig:trainingsetsize}, we investigate the relationship between the size of the training dataset and synthetic data generation performance. The x-axis is sorted based on the size of the training dataset in ascending order, as indicated in Table \ref{tab:data_description}, and the y-axis represents the average rank of distributional similarity metrics for each dataset. To enhance visualization clarity, we specifically select and present CW2, DistVAE, DistVAE (+CW), and our proposed model, CWDAE ($\pi=0.05$), which are the primary models of interest in this paper.

From Figure \ref{fig:trainingsetsize}, it becomes evident that CW2 (dashed orange line) consistently delivers inferior synthetic data generation performance across all tabular datasets. Additionally, DistVAE (dash-dot green line) exhibits significant variability in average rank, signifying that its synthetic data generation performance is notably dependent on the particular dataset being utilized.

In contrast, our proposed model, CWDAE (solid blue line), stands out by achieving the best performance, even for the smallest dataset, \texttt{loan} dataset, with a training dataset of around 4,000 samples. This highlights CWDAE's capability to effectively perform distributional learning even with a limited number of training samples. Furthermore, CWDAE's average rank remains relatively stable across different types of tabular datasets, demonstrating the robustness of its synthetic data generation performance, regardless of dataset variations. Additionally, when the training dataset size is sufficiently large, such as the \texttt{covtype} and \texttt{credit} datasets containing 45,000 samples, CWDAE outperforms other models, showcasing its superior performance under such conditions.

\subsubsection{Privacy preservability}

In assessing privacy preservation across various generative models, Table \ref{tab:privacy} provides a comprehensive comparison based on averaged metrics. Since DCR(S,S) is not a metric for privacy preservability, the `rank' column of Table \ref{tab:privacy} is averaged except for the rank of DCR(S,S). DCR(S,S) represents the distance between records within the synthetic dataset itself, and a higher score suggests the ability to create more diverse synthetic samples.
% (DCR(S,S) represents a metric for diversity of synthetic data sample). 

TVAE stands out as the top-performing model with the DCR(R,S), signifying that it maintains a significant distance between records in the real training dataset and those in the synthetic dataset. This indicates a high level of privacy preservation.

On the other hand, DistVAE achieves the highest metric score in DCR(S,S) in Table \ref{tab:privacy}. Another notable observation in the DCR(S,S) column of Table \ref{tab:privacy} is the remarkably low metric scores for CW2 and CW2 (+E) compared to other models, implying that CW2 and CW2 (+E) struggle to generate diverse synthetic samples. Therefore, as mentioned in Section \ref{sec:3.3}, models like DistVAE and CWDAE, which utilize quantile functions in decoder modeling and can create diverse synthetic samples based on inverse transform sampling method and various quantile levels, outperform models that construct the decoder using the Dirac delta function.

In terms of attribute disclosure metrics, CTAB-GAN achieves the best metric score for all nearest neighbor counts (1, 10, 100), positioning itself as a compelling model in the context of privacy preservation. Consequently, in the overall ranking of privacy preservability metrics, CTAB-GAN demonstrates the best performance (the average ranking is 1.2). This implies that CTAB-GAN could be a favorable choice in scenarios where the primary objective is preventing attribute disclosure. 

However, it's essential to note that the quality of synthetic data and the risk of privacy leakage are in a trade-off relationship. While CTAB-GAN exhibits the best privacy preservability in Table \ref{tab:privacy}, it also has the poorest synthetic data generation performance (it shows the worst average ranking of 13.7 in Table \ref{tab:metrics}). In contrast, our proposed CWDAE ($\pi=0.05$) achieves the best average ranking of 3.0 in Table \ref{tab:metrics}, demonstrating excellent synthetic data quality while maintaining a reasonably good privacy preservability (the average ranking is 7.0), indicating a balanced performance.

Notably, as the value of $\pi$ increases in CWDAE, both the DCR between the real training and synthetic datasets and the attribute disclosure metrics consistently increase. This suggests that the risk of privacy leakage can be controlled by adjusting $\pi$, where higher values of $\pi$ correspond to a higher level of privacy protection. As a result, by increasing $\pi$ up to 0.9, CWDAE shows the third-highest privacy preservability performance (average rank is 4.0) after CTAB-GAN.

\section{Conclusions}
\label{sec:5}

This paper introduces an innovative approach to generative model learning that aims to strike a balance between capturing both marginal and joint distributional characteristics. Our proposed model achieves this by utilizing a mixture measure (Definition \ref{def:mixcw}), which combines point masses on standard basis vectors with a normalized surface measure. This mixture measure is incorporated into the integral measure of the Cramer-Wold distance, resulting in a closed-form solution for the reconstruction loss, making it practical for implementation. Additionally, our model provides flexibility in adjusting the level of data privacy.

Our experiments demonstrate that the mixture Cramer-Wold distance outperforms existing measures such as the Cramer-Wold distance \cite{Knop2020GenerativeMW} or CRPS \cite{an2023distributional} when used to assess the dissimilarity between the generative model and the ground-truth distribution. This highlights the significance of incorporating marginal distributional information to enhance synthetic data generation performance.

Lastly, we want to emphasize that the mixture Cramer-Wold distribution can be applied to any generative model as a regularization term or for quantifying the discrepancy between any two high-dimensional probability distributions.

\section*{Acknowledgements}

The authors acknowledge the Urban Big data and AI Institute of the University of Seoul supercomputing resources (\url{http://ubai.uos.ac.kr}) made available for conducting the research reported in this paper.

\begin{table*}[t]
\caption{URL links for downloading six tabular datasets.}
  \centering
  \resizebox{0.85\linewidth}{!}{
  \begin{tabular}{c|cc}
    \toprule
    dataset & URL \\
    \midrule
    \texttt{covtype} & \url{https://www.kaggle.com/datasets/uciml/forest-cover-type-dataset} \\
    \texttt{credit} & \url{https://www.kaggle.com/c/home-credit-default-risk} \\
    \texttt{loan} & \url{https://www.kaggle.com/datasets/teertha/personal-loan-modeling} \\
    \texttt{adult} & \url{https://www.kaggle.com/datasets/uciml/adult-census-income} \\
    \texttt{cabs} & \url{https://www.kaggle.com/datasets/arashnic/taxi-pricing-with-mobility-analytics} \\
    \texttt{kings} & \url{https://www.kaggle.com/datasets/harlfoxem/housesalesprediction} \\
    \bottomrule
  \end{tabular}
}
\label{tab:urls}
\end{table*}

% \clearpage
\appendix
% \onecolumn
\section{Appendix}
\label{app:1}

\subsection{Objective Function of CWDAE}
\label{app:1.1}

\eqref{eq:logCWDAEpi} can be computed in the closed form, and the closed-form solution is written as
\bean
&& \cL(\theta,\phi) \\
&=& \pi \cdot \sum_{j=1}^D \alpha_j \left\| \frac{1}{n} \sum_{i=1}^n N(\bx_j^{(i)}, \gamma) - \frac{1}{n} \sum_{i=1}^n N(\hat{\bx}_j^{(i)}, \gamma) \right\|_2^2 \\
&& + (1-\pi) \cdot \int_{S_D} \left\| \mbox{sm}_\gamma(\nu^\top\bX) - \mbox{sm}_\gamma(\nu^\top\hat{\bX}) \right\|_2^2 d\sigma_D(\nu) \\
&& + \lambda \cdot \int_{S_d} \left\| \mbox{sm}_\gamma(\nu^\top\bZ) - \mbox{sm}_\gamma(\nu^\top\hat{\bZ}) \right\|_2^2 d\sigma_d(\nu) \\
&=& \pi \cdot \sum_{j=1}^D \alpha_j \Bigg\{ \frac{1}{n^2} \sum_{l=1}^n \sum_{k=1}^n \frac{1}{\sqrt{4\pi\gamma}} \exp \left( -\frac{1}{4\gamma} (\bx_j^{(l)} - \bx_j^{(k)})^2 \right) \\
&& + \frac{1}{n^2} \sum_{l=1}^n \sum_{k=1}^n \frac{1}{\sqrt{4\pi\gamma}} \exp \left( -\frac{1}{4\gamma} (\hat{\bx}_j^{(l)} - \hat{\bx}_j^{(k)})^2 \right) \\
&& - \frac{2}{n^2} \sum_{l=1}^n \sum_{k=1}^n \frac{1}{\sqrt{4\pi\gamma}} \exp \left( -\frac{1}{4\gamma} (\bx^{(l)}_j - \hat{\bx}_j^{(k)})^2 \right) \Bigg\} \\
&& + (1-\pi) \cdot \frac{1}{2n^2\sqrt{\pi\gamma}} \cdot \Bigg\{ \sum_{l=1}^n \sum_{k=1}^n \varphi_D \left( \frac{\|\bx^{(l)} - \bx^{(k)}\|^2}{4\gamma} \right) \\
&& + \sum_{l=1}^n \sum_{k=1}^n \varphi_D \left( \frac{\|\hat{\bx}^{(l)} - \hat{\bx}^{(k)}\|^2}{4\gamma} \right) \\
&& - 2 \sum_{l=1}^n \sum_{k=1}^n \varphi_D \left( \frac{\|\bx^{(l)} - \hat{\bx}^{(k)}\|^2}{4\gamma} \right) \Bigg\} \\
&& + \lambda \cdot \frac{1}{2n^2\sqrt{\pi\gamma}} \cdot \Bigg\{ \sum_{l=1}^n \sum_{k=1}^n \psi_d \left( \frac{\|\bz^{(l)} - \bz^{(k)}\|^2}{4\gamma} \right) \\
&& + \sum_{l=1}^n \sum_{k=1}^n \psi_d \left( \frac{\|\hat{\bz}^{(l)} - \hat{\bz}^{(k)}\|^2}{4\gamma} \right) \\
&& - 2 \sum_{l=1}^n \sum_{k=1}^n \psi_d \left( \frac{\|\bz^{(l)} - \hat{\bz}^{(k)}\|^2}{4\gamma} \right) \Bigg\},
\eean
where
\bean
\varphi_D(s) &\approx& \left( 1 + \frac{s}{2D - 3} \right)^{-1/2} \\
\psi_d(s) &=& \exp(-s/2) I_0\left(\frac{s}{2}\right),
\eean
$D \geq 20$, and we practically implement $\psi_d$ by applying the approximation of $I_0$ from p. 378 of \cite{Abramowitz1965HandbookOM} for $d=2$.

% \subsection{Dataset URLs}

% \subsection{Model Architecture}

\begin{table}[t]
\caption{The model architecture of CWDAE.}
\centering
  \resizebox{0.99\columnwidth}{!}{
  \begin{tabular}{c|c}
    \toprule
    encoder & decoder \\
    \midrule
    MLP(16) & MLP(16) \\
    ELU activation & ReLU activation \\
    MLP(8) & MLP(64) \\
    ELU activation & ReLU activation \\
    MLP($2 \times d$) & MLP($(M + 2) \times |I_c| + \sum_{j \in I_d} T_j|$) \\
    \bottomrule
  \end{tabular}
  }
\label{tab:arch}
\end{table}

\begin{table}[t]
\caption{The number of generative model parameters.}
\centering
  \resizebox{0.99\columnwidth}{!}{
  \begin{tabular}{lrrrrrr}
    \toprule
    Model & \texttt{covtype} & \texttt{credit} & \texttt{adult} & \texttt{loan} & \texttt{cabs} & \texttt{kings} \\
    \midrule
    CTAB-GAN & 12K & 13K & 14K & 5K & 12K & 15K \\
    CTGAN & 20K & 32K & 52K & 13K & 30K & 51K \\
    TVAE & 10K & 12K & 13K & 6K & 10K & 10K \\
    CW2 & 10.7K & 14.8K & 22.1K & 11.0K & 15.2K & 19.3K\\
    DistVAE & 9.4K & 11.5K & 11.5K & 5.9K & 8.8K & 14.5K\\
    \midrule
    CW2 (+Q) & 9.4K & 11.5K & 11.5K & 5.9K & 8.8K & 14.5K\\
    CW2 (+E) & 10.7K & 14.8K & 22.1K & 11.0K & 15.2K & 19.3K\\
    CW2 (+E,Q) & 9.4K & 11.5K & 11.5K & 5.9K & 8.8K & 14.5K\\
    CW2 (log, +E,Q) & 9.4K & 11.5K & 11.5K & 5.9K & 8.8K & 14.5K\\
    DistVAE (+CW) & 9.4K & 11.5K & 11.5K & 5.9K & 8.8K & 14.5K\\
    \midrule
    CWDAE ($\pi=0.05$) & 9.4K & 11.5K & 11.5K & 5.9K & 8.8K & 14.5K\\
    CWDAE ($\pi=0.1$) & 9.4K & 11.5K & 11.5K & 5.9K & 8.8K & 14.5K\\
    CWDAE ($\pi=0.5$) & 9.4K & 11.5K & 11.5K & 5.9K & 8.8K & 14.5K\\
    CWDAE ($\pi=0.9$) & 9.4K & 11.5K & 11.5K & 5.9K & 8.8K & 14.5K\\
    \bottomrule
  \end{tabular}
  }
\label{tab:num_params}
\end{table}

% \clearpage
\bibliographystyle{plain}
\bibliography{ref}

\end{document}